\documentclass[preprint,12pt]{elsarticle}

\usepackage{amssymb}
\usepackage{amsmath}
\usepackage{graphicx}
\usepackage{algorithm}
\usepackage{algpseudocode} 

\biboptions{sort&compress}

\usepackage[hidelinks]{hyperref}

\journal{Elsevier}

\begin{document}
	
\begin{frontmatter}



\title{Parameter-Free Structural-Diversity Message Passing for Graph Neural Networks}

\author[1]{Mingyue Kong}
\author[1]{Yinglong Zhang\corref{cor1}}
\ead{zyl1878@mnnu.edu.cn}
\ead{zhang_yinglong@126.com}  
\author[1]{Chengda Xu}
\author[1]{Xuewen Xia}
\author[1]{Xing Xu}

\address[1]{Minnan Normal University, No.\ 36 Xianqian Road, Zhangzhou 363000, Fujian, China}

\cortext[cor1]{Corresponding author.}

\begin{abstract}
	Graph Neural Networks (GNNs) have shown remarkable performance in structured data modeling tasks such as node classification. However, mainstream approaches generally rely on a large number of trainable parameters and fixed aggregation rules, making it difficult to adapt to graph data with strong structural heterogeneity and complex feature distributions. This often leads to over-smoothing of node representations and semantic degradation. To address these issues, this paper proposes a parameter-free graph neural network framework based on structural diversity, namely SDGNN (Structural-Diversity Graph Neural Network). The framework is inspired by structural diversity theory and designs a unified structural-diversity message passing mechanism that simultaneously captures the heterogeneity of neighborhood structures and the stability of feature semantics, without introducing additional trainable parameters. Unlike traditional parameterized methods, SDGNN does not rely on complex model training, but instead leverages complementary modeling from both structure-driven and feature-driven perspectives, thereby effectively improving adaptability across datasets and scenarios. Experimental results show that on eight public benchmark datasets and an interdisciplinary PubMed citation network, SDGNN consistently outperforms mainstream GNNs under challenging conditions such as low supervision, class imbalance, and cross-domain transfer. This work provides a new theoretical perspective and general approach for the design of parameter-free graph neural networks, and further validates the importance of structural diversity as a core signal in graph representation learning. To facilitate reproducibility and further research, the full implementation of SDGNN has been released at: \url{https://github.com/mingyue15694/SGDNN/tree/main}
\end{abstract}





\end{frontmatter}



\section{Introduction}
\label{sec:intro}

With the increasing scale and structural complexity of complex networks such as social media, academic knowledge graphs, and biomedical networks, how to efficiently obtain high‑quality node representations has become a central problem in tasks such as information retrieval \cite{ref1}, recommender systems \cite{ref2}, and network analysis \cite{ref3}. In recent years, Graph Neural Networks (GNNs), by integrating node features with topological structures, have achieved widespread success in tasks such as node classification \cite{ref4}, link prediction \cite{ref5}, and community detection \cite{ref6}. However, when applied to large‑scale complex graphs in the real world, they often face multiple challenges.

First, mainstream GNNs typically rely on a large number of trainable parameters. When node feature dimensions are high, distributions are sparse, and labeled data is limited, such designs not only struggle to generalize effectively but are also prone to overfitting \cite{ref43}. Second, fixed convolutional or attention mechanisms aggregate all neighbors in the same way, ignoring potential structural and semantic differences within neighborhoods. After multiple stacked layers, such homogenized processing tends to cause over‑smoothing \cite{ref44}, leading to representations of different categories of nodes becoming increasingly indistinguishable.

Moreover, existing methods generally lack explicit modeling of neighborhood {structural diversity}, which is critical in complex scenarios such as cross-community propagation and cross-domain information fusion. This deficiency is particularly evident in practice. For example, in academic collaboration networks, a researcher may collaborate closely with scholars in computer science while also maintaining ties with the bioinformatics community. Standard message passing mechanisms (e.g., GraphSAGE~\cite{ref23} or GCN~\cite{ref21}) directly mix features from both communities, diluting the key signal of cross-domain collaboration and resulting in blurred, averaged semantics. How to identify independent branches within a neighborhood (e.g., computer science vs.\ bioinformatics communities) while preserving semantic differences remains an open question in GNN message passing research.

Structural diversity, inspired by the theories of weak ties \cite{ref16} and structural holes \cite{ref17}, offers a way to capture this neighborhood complexity \cite{ref7}. By counting the number of independent groups in a node’s neighborhood, it measures the breadth of heterogeneous information accessible to the node. Prior studies have shown that nodes with high structural diversity play critical roles in information diffusion, cross‑community collaboration, and acting as bridges in networks \cite{ref8,ref9}. Although this concept has been well validated in fields such as social network analysis and information propagation modeling, there is still a lack of GNN designs explicitly grounded in structural diversity theory.

To address this, we propose SDGNN, a parameter‑free graph neural network framework that incorporates structural diversity awareness. SDGNN aims to balance expressive power, modeling efficiency, and structural generalization under a completely parameter‑free setting. At its core lies the Structural‑Diversity Message Passing (SDMP) mechanism we design, which combines structural partitioning with feature clustering to enable multi‑level neighborhood segmentation and information integration, while adapting to structural–semantic heterogeneity across different layers.

\begin{figure}[ht]
	\centering
	\includegraphics[width=0.85\linewidth]{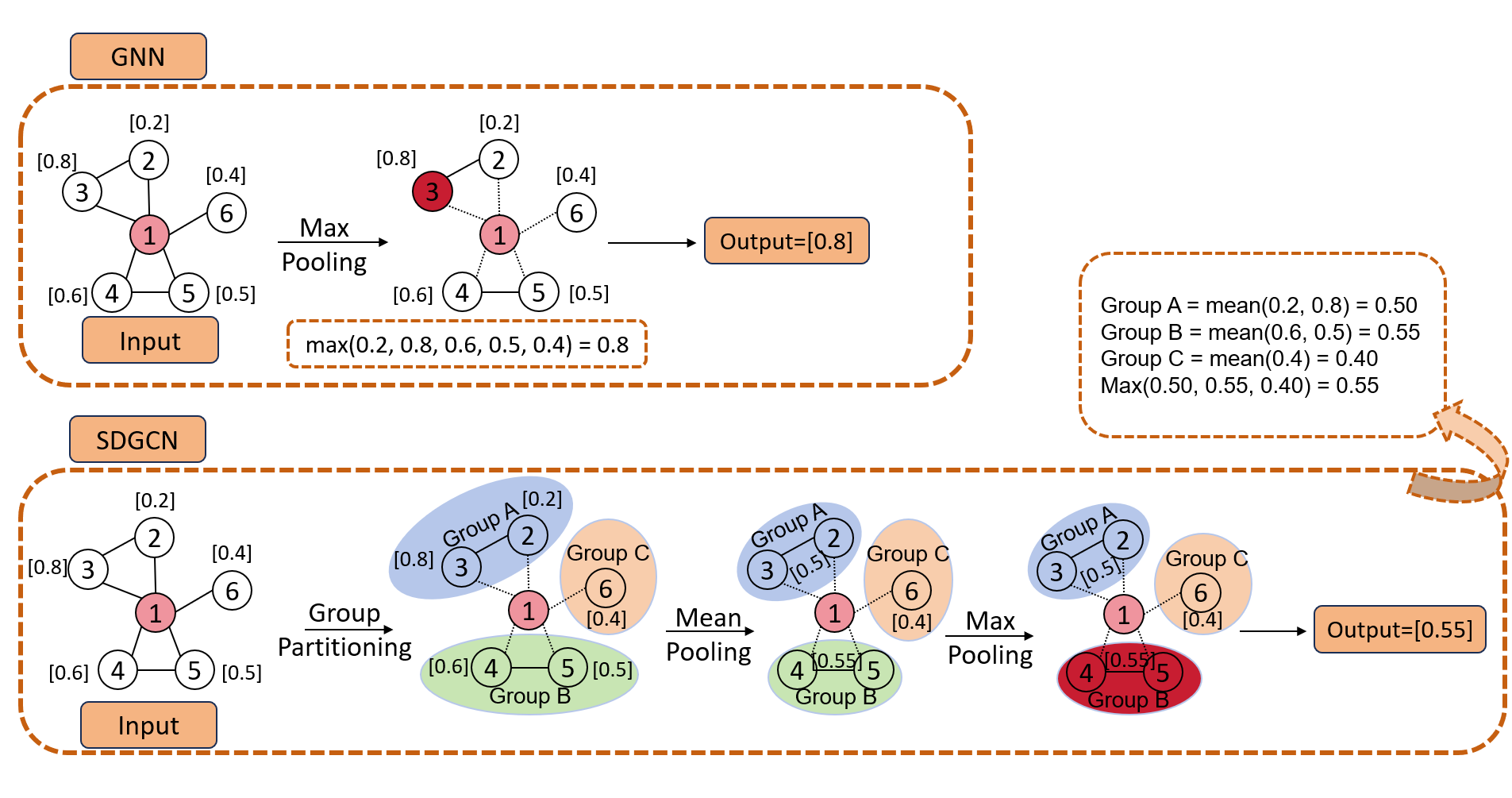} 
	\caption{Comparison of aggregation mechanisms between GNN and SDGNN.}
	\label{fig:fig1}
\end{figure}

Figure~\ref{fig:fig1} compares a simplified standard GNN with SDGNN. As shown, Node~2 and Node~3 are direct neighbors and belong to the same branch structure, but their values differ greatly. Node~3’s feature value (0.8) is much higher than that of other neighbors, which may be an outlier introduced due to modeling error. In the figure, the standard GNN treats all neighbors equally during aggregation, and after applying max pooling, the output is directly dominated by this outlier (Output $=0.80$). In contrast, inspired by structural diversity theory, SDGNN groups Node~1’s neighborhood based on connectivity (Groups A/B/C). The feature update of Node~1 is no longer influenced by a single direct neighbor, but rather by the three groups A, B, and C. Specifically, within each group, a group consensus is adopted to mitigate the impact of local anomalies (Group~A: mean $=0.50$; Group~B: mean $=0.55$; Group~C: mean $=0.40$). Then, across groups, max pooling is applied, and the final output is stably dominated by Group~B (Output $=0.55$). This process effectively suppresses interference from a single outlier while preserving semantic diversity across branches. Note that this example is a simplified illustration, omitting common operations such as feature concatenation or linear transformation, to highlight the core differences in the aggregation mechanism.

The key to the structural‑diversity message passing mechanism lies in neighborhood group partitioning. To this end, SDGNN designs three complementary group‑partition strategies: a DBSCAN‑based \cite{ref11} density clustering strategy for highly heterogeneous structures; a structure‑guided dynamic aggregation mechanism that balances efficiency and expressiveness; and a fully feature‑driven non‑iterative pseudo‑aggregation mechanism that emphasizes semantic consistency and computational efficiency. Multi‑layer node representations obtained under these strategies are then fused through the Jumping Knowledge (JK) mechanism \cite{ref14}, and passed to a lightweight multilayer perceptron (MLP) \cite{ref15} to generate the final outputs for downstream tasks such as node classification and cross‑domain recognition. On this basis, to further enhance the model’s ability to capture long‑range semantic dependencies and global structural features, we propose a normalization‑propagation‑based structural enhancement mechanism as a global modeling supplement to the SDGNN framework.

Message passing in Graph Neural Networks can, in a sense, be regarded as a form of social contagion. We attempt to establish a bridge connecting sociological theories (weak ties, structural holes, and structural diversity) to graph neural networks, even though we have not elaborated much on the relevant connections. As an exploratory endeavor, we hope this work will inspire researchers to construct interpretable, superior, and more innovative graph neural networks based on classical sociological theories. The main contributions of this paper are summarized as follows:

\begin{enumerate}
	\item We propose a parameter-free GNN framework driven by structural diversity. This framework establishes a unified structural-diversity message passing paradigm that explicitly captures neighborhood structural heterogeneity, enabling efficient and interpretable node representation learning.
	
	\item We systematically introduce three complementary group-partition strategies (structure-driven, feature-driven, and hybrid), balancing efficiency, adaptability, and expressiveness. The structure-driven partition, combined with a normalization-propagation-based global structural enhancement module, is further proposed to capture global structural and semantic features.
	
	\item We conduct comprehensive experiments on eight public benchmark datasets and an interdisciplinary PubMed citation network, demonstrating the superiority of SDGNN in structurally complex scenarios. Moreover, in interdisciplinary tasks, we reveal the complementary roles of different cross-group fusion strategies, highlighting the model’s stability and generalization ability across diverse scenarios.
\end{enumerate}

\section{Related Work}
\label{sec:related}

\subsection{Structural Diversity}
\label{subsec:sd}

Structural diversity originates from social network analysis and measures the breadth of an individual's connections across communities and the heterogeneity of their information sources. In \emph{The Strength of Weak Ties}, Granovetter showed that weak ties can bridge community boundaries and facilitate the diffusion of heterogeneous information \cite{ref16}. Building on this, the theory of structural holes further reveals that nodes located at structural holes can span multiple disconnected groups, efficiently integrate multi‑source information, and act as bridges in networks \cite{ref17}.

Ugander et al.\ first provided a quantitative measure of structural diversity on large‑scale Facebook networks by defining it as the number of independent connected components in a node’s neighborhood; they verified its significant correlation with decision‑making, innovation, and diffusion efficiency \cite{ref7}. This concept has broad practical value: in social network analysis, nodes with high structural diversity often serve as cross‑community information hubs, aiding the identification of key information nodes \cite{ref18}; in information diffusion research, such nodes tend to become diffusion cores due to their coverage of multiple propagation paths \cite{ref19}; and in complex network evaluation, structural diversity reveals node multifunctionality and fine‑grained topological characteristics \cite{ref20}. Recently, the Citation Structural Diversity (CSD) index integrates structural relationships with semantic features for literature and journal evaluation, extending structural diversity to scientometrics \cite{ref45}.

\subsection{Graph Neural Networks}
\label{subsec:gnn}

Graph Neural Networks (GNNs) integrate local topology and feature information through message passing and feature aggregation, and have become the mainstream methods for handling graph-structured data \cite{ref42}. Typical models include GCN \cite{ref21}, which is based on spectral convolution and smooths neighbor features with weights, and GAT \cite{ref22}, which assigns differentiated weights to neighbors via attention mechanisms, thereby enhancing information filtering. However, the aggregation rules of these methods are fixed. When neighborhoods contain diverse sub-communities or strong heterogeneity, they fail to effectively distinguish different semantic sources and are prone to over-smoothing of representations.

To improve adaptability to structural differences, various extensions have been proposed in aggregation strategies. GraphSAGE \cite{ref23} improves scalability on large-scale graphs through neighbor sampling, but simple aggregation may obscure neighborhood differences. GIN \cite{ref24} enhances the ability to distinguish non-isomorphic nodes, but still suffers from over-smoothing and parameter dependence in deeper layers. JK-Net \cite{ref14} employs jumping connections to retain multi-layer semantic features, alleviating deep-layer degradation to some extent, but its ability to model neighborhood diversity remains limited. SAT \cite{ref25} performs structure-adaptive aggregation by adjusting strategies according to neighborhood patterns, but it relies on trainable parameters and is prone to overfitting under low-supervision conditions. SGCN \cite{ref13} improves efficiency through simplified multi-hop propagation, but lacks explicit modeling of neighborhood structures, thereby limiting representation discriminability.

Meanwhile, some studies attempt to introduce community partitioning or sparse selection mechanisms to enhance structural awareness. For example, TANGNN \cite{ref26} selects the most contributive neighbors through Top-m attention, suitable for handling heterogeneous graphs with sparsity or long-range dependencies, but incurs high computational and storage costs on large-scale graphs. ComBSAGE \cite{ref10} partitions neighbor-induced subgraphs into connected components and performs two-stage aggregation, improving the modeling of local community structures, but lacks the utilization of global structural signals and relies on fixed partitioning strategies. AgentNet \cite{ref38} employs intelligent agents to walk through graphs and collect information, enabling the recognition of complex substructures and showing strong expressive ability in graph-level tasks. More recently, RSEA-MVGNN \cite{ref41} introduced a structural enhancement module within a multi-view framework, dynamically adjusting aggregation weights by measuring structural differences across views, providing new insights into the development of structure-sensitive GNNs.

In addition, some studies have extended graph representation learning from the perspectives of structural signals and adversarial mechanisms. Ma et al.\ \cite{ref39} proposed CoS-GNN, which introduces rich structural features into graphs to enhance message passing and improve performance in graph-level tasks. Yu et al.\ \cite{ref40} proposed CS-GNN, which leverages attention mechanisms to capture structural context and model structural patterns across nodes. Zhang et al.\ \cite{ref46} introduced a motif-based structural attention mechanism combined with dual negative sampling for adversarial learning, effectively improving the discriminability and robustness of graph representations. Yang et al.\ \cite{ref47} proposed an adversarial learning framework that preserves node similarity, ensuring semantic consistency under adversarial perturbations and enhancing the stability and generalization of graph representations.

Unlike the aforementioned studies, this paper, inspired by sociological theory, proposes SDGNN, a parameter-free graph neural network framework driven by structural diversity. It captures multi-source semantic features at both local and global levels in a staged manner, achieving efficient modeling and robust generalization of node representations.

\section{Basic Definitions}
\label{sec:definitions}

\textbf{Definition 1 (Neighborhood Set).} 
Let an undirected graph $G=(V,E)$, where $V$ is the node set, $|V|=n$ denotes the number of nodes, and $E \subseteq V \times V$ is the edge set. 
For any node $v \in V$, its neighborhood set is defined as:
\begin{equation}
	N(v) = \{ u \in V \mid (v,u) \in E \}.
	\label{eq:neighborhood}
\end{equation}
The original feature vector of node $v$ is denoted as $x_v$.

\medskip
\textbf{Definition 2 (Node Representation).} 
At layer $l$, the feature representation of node $v$ is denoted as $h_v^{(l)}$. 
For the initial layer (input layer), we have:
\begin{equation}
	h_v^{(0)} = x_v.
	\label{eq:node-representation}
\end{equation}

\medskip
\textbf{Definition 3 (Structural Diversity).} 
In graph $G=(V,E)$, for any node $v \in V$, let its one-hop neighborhood be $N(v)$. 
Consider the induced subgraph $G[N(v)]$, and extract all connected components. 
The set of these connected components is denoted as:
\begin{equation}
	C_v = \{ C_{v,1}, C_{v,2}, \ldots, C_{v,S_v} \},
	\label{eq:components}
\end{equation}
where $C_{v,s}$ denotes the node set of the $s$-th connected component, such that
\begin{equation}
	\bigcup_{s=1}^{S_v} C_{v,s} = N(v),  
	C_{v,i} \cap C_{v,j} = \varnothing \;\; \forall i \neq j.
	\label{eq:component-constraints}
\end{equation}
Then, the structural diversity of node $v$ is defined as the number of connected components in its neighborhood:
\begin{equation}
	SD_v = S_v.
	\label{eq:sd}
\end{equation}
This metric reflects the degree of structural independence among the neighbors connected to a node, characterizing the structural heterogeneity of its information sources.

\medskip
\textbf{Definition 4 (General Structural Diversity).} 
In the generalized case, structural diversity is not limited to connected components but allows neighbors to be partitioned into different groups based on any similarity criteria, such as structural or feature-based measures. 
Formally, for node $v \in V$, its neighborhood set $N(v)$ is partitioned into several disjoint groups according to a chosen similarity measure. 
The set of groups is denoted as:
\begin{equation}
	C_v = \{ C_{v,1}, C_{v,2}, \ldots, C_{v,S_v} \},
	\label{eq:general-partition}
\end{equation}
where
\begin{equation}
	C_{v,i} \cap C_{v,j} = \varnothing \; \forall i \neq j, \;
	\bigcup_{s=1}^{S_v} C_{v,s} = N(v).
	\label{eq:general-constraints}
\end{equation}
Then, the generalized structural diversity of node $v$ is defined as:
\begin{equation}
	GSD_v = S_v.
	\label{eq:gsd}
\end{equation}
Unlike Definition~\ref{eq:components}, which relies solely on graph structure, generalized structural diversity allows group construction based on any neighborhood partitioning strategy (e.g., density-based clustering, semantic similarity), thereby providing a more comprehensive characterization of the structural complexity and semantic heterogeneity of a node’s environment.

\section{Structural-Diversity Message Passing Graph Neural Network}
\label{sec:sdgnn}

GNNs have become mainstream techniques for graph data modeling due to their ability to effectively integrate graph structure with node features. However, most existing GNN models rely heavily on a large number of trainable parameters and adopt fixed convolutional or weighted aggregation rules. When dealing with graph data characterized by strong structural heterogeneity and complex feature distributions, they often suffer from over-smoothed representations, semantic degradation, and limited generalization capability.

To overcome these challenges, this paper proposes a novel parameter-free graph neural network framework, SDGNN. At its core lies the SDMP mechanism, which explicitly models the diverse interactions between a node and its neighbor groups. Without introducing any trainable aggregation parameters, SDMP enables a flexible, unified, and expressive feature aggregation process.

\begin{figure}[t]
	\centering
	\includegraphics[width=0.85\linewidth]{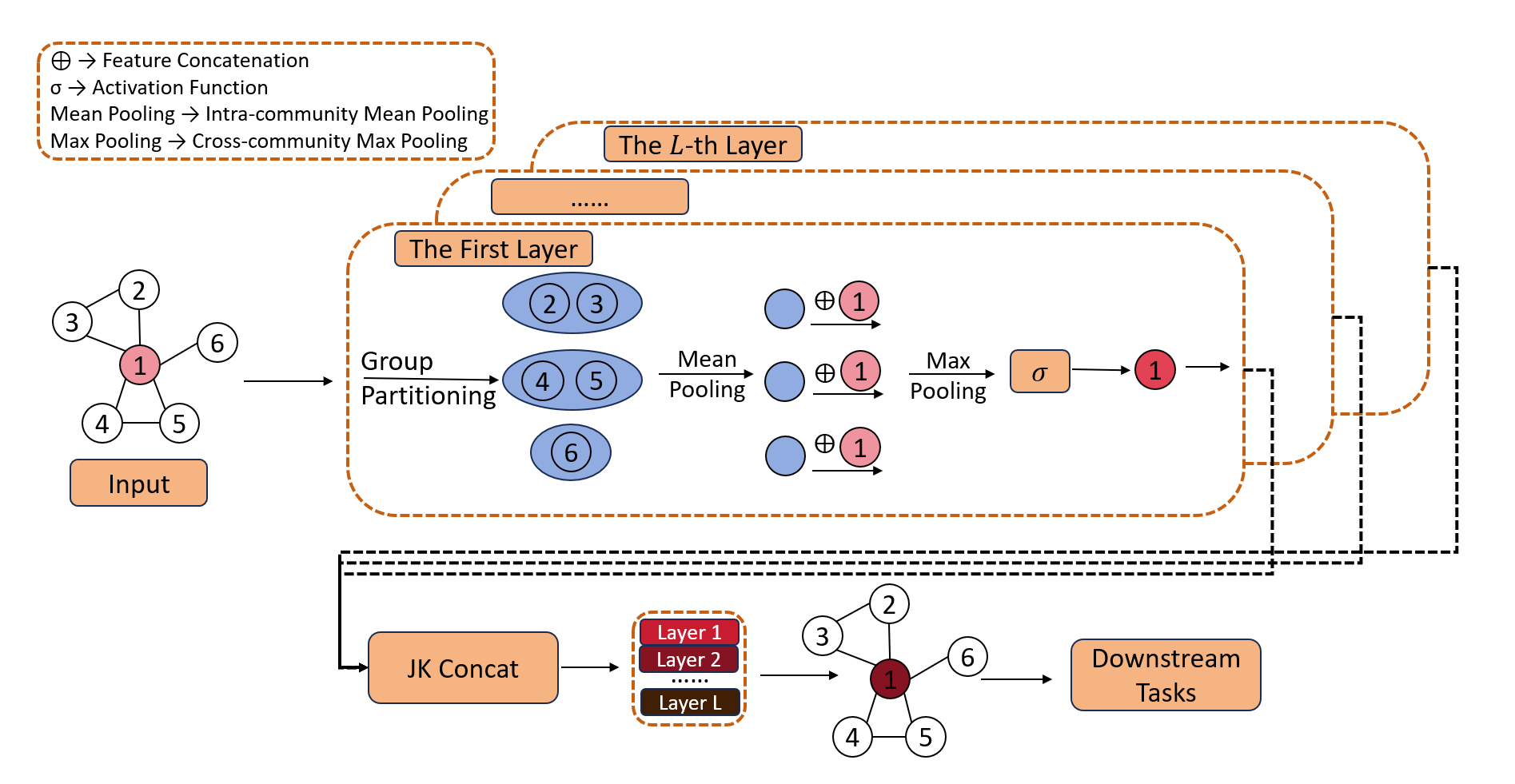} 
	\caption{Overall architecture of the SDGNN model.}
	\label{fig:fig2}
\end{figure}

Figure~\ref{fig:fig2} illustrates the overall computational flow of SDGNN. Taking a target node $v$ (red node) in the input graph as an example, the model first identifies its neighborhood $N(v)$ (white nodes), and then, at each layer, partitions these neighbors into several groups with structural or semantic consistency based on predefined strategies. This partitioning reflects the fundamental idea of SDMP, which uses generalized structural diversity as the modeling signal: the semantic information of a node is no longer derived from individual neighbors, but from the structural consensus of the groups to which those neighbors belong. For each subgroup, mean pooling is applied to encode local semantic features. These group features are then concatenated with the representation of the central node, forming semantic interaction information between the node and its groups as input to the downstream module. To select the most informative aggregation from multiple groups, the model applies max pooling across all concatenated results, followed by a nonlinear activation function (e.g., ReLU) to complete the representation update at that layer. 

This process is repeated in a multi-layer stacked structure, with different layers adopting different neighborhood partition strategies to capture multi-scale variations in semantic expression. Finally, SDGNN fuses node representations across layers through a Jumping Knowledge mechanism \cite{ref14}, thereby constructing globally structure-aware node embeddings for downstream tasks such as classification or interdisciplinary prediction.

Throughout the entire model, all feature update operations are carried out under the unified paradigm of SDMP, with different specific SDMPs corresponding to different neighbor group partitioning strategies. The following subsections will first introduce the Structural-Diversity Message Passing mechanism (Section~\ref{sec:sdmp}) and its implementation under three parameter-free group partition strategies (Sections~\ref{subsec:sdmp-density}--\ref{subsec:sdmp-feature}). Subsequently, in Section~\ref{subsec:theory}, we further provide a theoretical analysis of SDMP, focusing on its permutation invariance property to ensure stability and robustness against different neighborhood orderings.

\subsection{Structural-Diversity Message Passing Mechanism}
\label{sec:sdmp}

To systematically characterize semantic relations of nodes within complex neighborhoods, we propose a Structural-Diversity Message Passing mechanism as the general paradigm, which serves as the core computational primitive of the SDGNN framework. 

This mechanism is inspired by the theories of weak ties \cite{ref16} and structural holes \cite{ref17} in social network studies, emphasizing the critical role of group structures—rather than individual neighbors—in node semantic modeling. We argue that the semantic role of a node depends not only on the content of its neighbors but also on how these neighbors are organized into different structural or semantic groups. Compared to the sheer number of neighbors, the way neighborhoods are partitioned and the diversity of groups reveal the complexity of information sources and the heterogeneity of interaction contexts more effectively. 

Guided by this idea, SDMP takes generalized structural diversity as the modeling driver (see Equation~\ref{eq:sdmp-agg}), partitioning a node’s neighborhood into several structurally or semantically coherent groups, and performing local statistical feature extraction and integration around these groups. Its core aggregation expression can be formalized as:
\begin{equation}
	h_v^{(l)} = \sigma \Bigg( 
	\max_{C \in \mathcal{C}_v^{(l)}} 
	\Big[ \, h_v^{(l-1)} \;\; \oplus \;\; 
	\frac{1}{|C|}\sum_{u \in C} h_u^{(l-1)} \, \Big] 
	\Bigg),
	\label{eq:sdmp-agg}
\end{equation}
where $h_v^{(l)}$ denotes the representation of node $v$ at layer $l$; $\mathcal{C}_v^{(l)}=\{C_{v,1},C_{v,2},\ldots,\\C_{v,S_v}\}$ is the set composed of several groups partitioned from the neighbor set N(v) of node v at the $l$-th layer, with $S_v$ groups in total; each subgroup $C \in \mathcal{C}_v^{(l)}$, where $C \subseteq N(v)$, is formed based on a certain similarity measure; $\oplus$ denotes the feature concatenation operation; $\frac{1}{|C|}\sum_{u \in C} h_u^{(l-1)}$ represents mean pooling within each group; $\max(\cdot)$ denotes dimension-wise max pooling across groups, selecting the most informative group; and $\sigma(\cdot)$ is the activation function, with ReLU as the default.

This formula depicts a widely adaptable parameter-free aggregation framework. It emphasizes modeling neighbors at the \emph{group level}, which effectively avoids over-averaging, preserves semantic discriminability, and naturally satisfies the properties of being parameter-free, symmetric, and extensible.

\begin{figure}[t]
	\centering
	\includegraphics[width=0.85\linewidth]{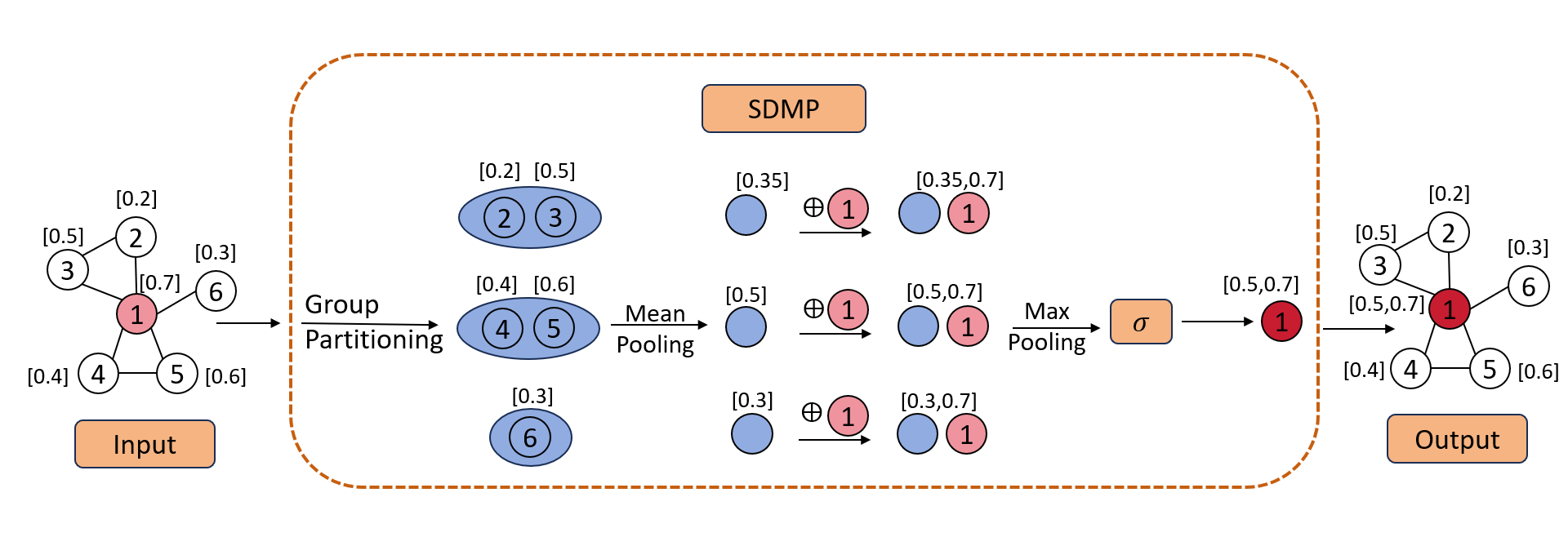} 
	\caption{Logical framework of the Structural-Diversity Message Passing (SDMP) mechanism.}
	\label{fig:fig3}
\end{figure}

To facilitate understanding of the computational process in Equation~\ref{eq:sdmp-agg}, Figure~\ref{fig:fig3} provides a concrete example. 
Take node $v=1$ as an example, with neighborhood $N(1)=\{2,3,4,5,6\}$. 
Under a certain neighborhood partitioning strategy, its neighbors are divided into three groups:
\begin{equation}
	\mathcal{C}_1 = \{ C_{1,1}=\{2,3\}, \; C_{1,2}=\{4,5\}, \; C_{1,3}=\{6\} \}.
\end{equation}

The feature representations of the nodes at the previous layer are given as:
\begin{equation}
	\begin{aligned}
		h_1^{(l-1)}&=[0.7], \; h_2^{(l-1)}=[0.2], \; h_3^{(l-1)}=[0.5],\\
		h_4^{(l-1)}&=[0.4], \; h_5^{(l-1)}=[0.6], \; h_6^{(l-1)}=[0.3].
	\end{aligned}
\end{equation}

According to the SDMP mechanism, mean pooling is performed within each structural community $C_{v,s} \subseteq N(v)$ to compute their statistical features:
\begin{align}
	m_{1,1} &= \tfrac{1}{2}(x_2 + x_3) = \tfrac{1}{2}(0.2 + 0.5) = [0.35], \\
	m_{1,2} &= \tfrac{1}{2}(x_4 + x_5) = \tfrac{1}{2}(0.4 + 0.6) = [0.5], \\
	m_{1,3} &= x_6 = [0.3].
\end{align}

Then, each community’s mean feature is concatenated with the center node’s feature $x_1$ (using $\oplus$ to denote concatenation):
\begin{align}
	z_{1,1} &= x_1 \oplus m_{1,1} = [0.7, \; 0.35], \\
	z_{1,2} &= [0.7, \; 0.5], \\
	z_{1,3} &= [0.7, \; 0.3].
\end{align}

Next, cross-group max pooling is applied, selecting the maximum value in each dimension to obtain the aggregated vector representing the most significant group contribution:
\begin{equation}
	h_1^{(l)} = \sigma\!\left(\max\{ z_{1,1}, z_{1,2}, z_{1,3} \}\right) = [0.7, \; 0.5].
\end{equation}

The general computational process of SDMP is shown in Algorithm~\ref{alg:sdmp}. 
It is important to emphasize that the differences among all strategies lie solely in the way neighbor groups are partitioned. 
In contrast, the aggregation expression and computational logic remain consistent, ensuring the universality and composability of the entire framework.

\begin{algorithm}[t]
	\caption{Generalized SDMP Aggregation}
	\label{alg:sdmp}
	\begin{algorithmic}[1]
		\Require Node $v \in V$; neighbor set $N(v)$; previous-layer features $\{h_u^{(l-1)} \mid u \in N(v)\}$; group partition strategy $\textsc{GroupPartition}(\cdot)$; activation function $\sigma$.
		\Ensure Updated node representation $h_v^{(l)}$.
		\If{$N(v) = \emptyset$}
		\State $h_v^{(l)} \gets \sigma\!\big([\,h_v^{(l-1)} \oplus h_v^{(l-1)}\,]\big)$
		\Else
		\State $\mathcal{C}_v^{(l)} \gets \textsc{GroupPartition}\big(N(v)\big)$
		\State $Z \gets \varnothing$
		\For{each group $C \in \mathcal{C}_v^{(l)}$}
		\State $m_C \gets \mathrm{mean}\big(\{\,h_u^{(l-1)} \mid u \in C\,\}\big)$
		\State $z_C \gets [\,h_v^{(l-1)} \oplus m_C\,]$
		\State $Z \gets Z \cup \{ z_C \}$
		\EndFor
		\State $h_v^{(l)} \gets \sigma\big( \max(Z) \big)$
		\EndIf
		\Return $h_v^{(l)}$
	\end{algorithmic}
\end{algorithm}

\subsection{Parameter-Free Group Partitioning Strategies in SDMP}
\label{subsec:pf-group-partition}

In the SDGNN framework, the update of node representations fully relies on the unified feature aggregation paradigm defined by the SDMP mechanism. This paradigm takes groups as the minimal interaction units, partitioning the neighborhood of each layer into several internally consistent subsets, and performing feature aggregation within each group. In essence, the way groups are partitioned determines the granularity at which the model understands neighborhood semantic structures, making it a key bridge between local perception and global semantic modeling. Therefore, although the computational process of SDMP remains consistent across all layers, the choice of group partitioning strategy directly affects the effectiveness of representation learning. To adapt to different graph structures and task requirements, we design three complementary parameter-free group partitioning strategies, which focus respectively on structure-driven, feature-driven, and hybrid modeling approaches. All these strategies are designed without introducing any trainable parameters, thereby ensuring modeling efficiency and generalization ability while maintaining expressive power.

\subsubsection{Group Partitioning Strategy Based on Structural Division and Density Clustering}
\label{subsec:sdmp-density}

This strategy combines structure-driven and feature-driven collaborative modeling within the SDMP framework. It first performs community partitioning using topological structure and then identifies semantically consistent groups in the feature space using DBSCAN, thus balancing structural heterogeneity with semantic consistency. In networks with significant structural differences, relying solely on feature similarity may fail to accurately capture neighborhood patterns, whereas this method can preserve both topologically independent branches and semantically tight clusters. It is particularly suitable for complex graphs where a node's neighborhood contains multiple dense substructures, such as cross-community collaboration networks or multi-domain citation networks.

Specifically, at the first layer, the model extracts all connected components from the induced subgraph $G[N(v)]$ of the target node’s neighborhood, treating them as local structural communities to explicitly model structural diversity. At the second layer and beyond, the DBSCAN clustering algorithm is introduced, using the previous layer's node representations to partition neighbors in the feature space and extract semantically consistent substructures, uncovering potential local semantic patterns.

\paragraph{First layer (structure-driven stage)}\label{subsec:feature-stage}

If $N(v)=\emptyset$, then its output representation is set as the concatenation of its own features:
\begin{equation}
	h_v^{1}=\sigma\!\left(\left[h_v^{0}\oplus h_v^{0}\right]\right),
\end{equation}
where $h_v^{0}$ is the initial feature of node $v$.

If $N(v)\neq\emptyset$, we extract all connected components from its induced subgraph $G[N(v)]$, denoted as
\begin{equation}
	C_v=\{C_{v,1},C_{v,2},\ldots,C_{v,S_v}\},
\end{equation}
where $\bigcup_{s=1}^{S_v}C_{v,s}=N(v)$ and all communities are disjoint. For each structural community, we compute the mean feature of its members and concatenate this with the representation of the central node. Across all such community-level vectors, max pooling followed by activation is applied to obtain the final update. This procedure does not introduce any trainable parameters and directly instantiates the SDMP aggregation logic described in Eq.(9): local statistics are extracted within groups, then combined across groups to preserve structural diversity while mitigating the effect of noisy neighbors.

\paragraph{Higher layers ($\ell \ge 2$, feature-driven stage)}\label{subsec:feature-stage}

DBSCAN is employed as a feature-driven clustering method to partition neighbors in the feature space, thereby obtaining finer-grained local semantic structures. Given the input representations from the previous layer $h_v^{l-1}$ and the neighbor set
\begin{equation}
	H_v^{l-1}=\{\,h_u^{l-1}\mid u\in N(v)\,\},
\end{equation}
we design the following three cases based on neighborhood size.

\noindent\textit{If $N(v)=\emptyset$:} the node features are concatenated and repeated as output:
\begin{equation}
	h_v^{l}=\sigma\!\left(\big[h_v^{l-1}\oplus h_v^{l-1}\big]\right).
\end{equation}

\noindent\textit{If $\lvert N(v)\rvert\le 2$:} mean pooling is applied directly to the neighbor features:
\begin{equation}
	h_v^{l}=\sigma\!\left[\,h_v^{l-1}\oplus \frac{1}{\lvert N(v)\rvert}\sum_{u\in N(v)} h_u^{l-1}\right].
\end{equation}

\noindent\textit{If $\lvert N(v)\rvert>2$:} DBSCAN clustering is applied to the neighbor feature set $H_v^{l-1}$, resulting in cluster aggregation $\{C_1,C_2,\ldots,C_M\}$. If all neighbors are treated as noise, the operation simply degenerates to mean pooling. Otherwise, each valid cluster is represented by the average of its members, which is then concatenated with the central node’s feature. Finally, all cluster-level vectors are integrated through cross-group max pooling and nonlinear activation to produce the updated node representation.

This procedure is essentially another instantiation of the SDMP aggregation rule in Eq.(9): feature statistics are computed within each cluster, then summarized across clusters to preserve semantic diversity. The strategy thus maintains the parameter-free property while adaptively capturing semantically consistent substructures. However, due to DBSCAN’s reliance on density estimation and distance computations, its scalability in high-dimensional spaces can be limited.

\subsubsection{Structure-Guided Dynamic Group Partitioning Strategy}
\label{subsec:sdmp-feature}

To overcome the high computational cost of DBSCAN in high-dimensional feature spaces, this strategy introduces a structure-guided dynamic aggregation mechanism. It adaptively determines the number of clusters based on the result of the first-layer structural partitioning and performs the clustering assignment in the feature space in a single step, without the need for iterative updates. This approach significantly reduces computational costs while preserving structural sensitivity, making it especially suitable for large-scale networks or tasks that require high responsiveness. It is a parameter-free alternative that strikes a balance between efficiency and expressive power.

This mechanism continues the structural partitioning idea based on connected components from the neighbor-induced subgraph in the first layer and uses it as a guiding signal for dynamically determining the number of clusters in subsequent layers. In layer $l\geq 2$, the model no longer explicitly relies on graph structure but instead partitions the neighbors based on the previous layer’s node representations in the feature space, enabling flexible and efficient local semantic aggregation.

Specifically, for any node $v\in V$, the structural partitioning in the first layer provides a set of connected subgraphs for its neighborhood:
\begin{equation}
	C_v=\{C_{v,1},C_{v,2},\ldots,C_{v,K_v}\},
\end{equation}
where $K_v$ represents the number of groups into which node $v$’s neighbors are partitioned under structure-guided clustering. This value is directly used as the number of dynamic clusters for the current layer. For each group $C_{v,k}\subseteq N(v)$, the mean feature of its members from the previous layer is computed as the initial center:
\begin{equation}
	c_k=\frac{1}{\lvert C_{v,k}\rvert}\sum_{u\in C_{v,k}} h_u^{(l-1)},\quad k=1,2,\ldots,K_v.
\end{equation}
Then, for each neighbor node $u\in N(v)$, it is assigned to the nearest cluster based on the Euclidean distance to each center $c_k$, yielding a new non-overlapping set of neighbor clusters:
\begin{equation}
	\mathcal{C}_v^{(l)}=\{C_1,C_2,\ldots,C_{K_v}\}.
\end{equation} 

Once the new clusters are formed, the feature update step directly follows the SDMP aggregation rule in Eq.(9): each cluster’s mean feature is concatenated with the central node’s representation, and the final node embedding is obtained via cross-group max pooling and nonlinear activation. If the neighborhood is empty, the update degenerates to self-concatenation. This strategy ensures that the number of clusters adapts to the structural partition while the aggregation remains parameter-free and consistent with the unified SDMP framework, effectively balancing efficiency and semantic expressiveness.

This method achieves key breakthroughs in three areas. First, the number of clusters is no longer fixed, but is adaptively determined by the initial structure, enhancing the model’s adaptability to complex neighborhood structures. Second, the partitioning process does not rely on any iteration, requiring only a one-time initialization and nearest-neighbor assignment, significantly reducing computational complexity. Third, the clustering logic is entirely driven by feature distribution, freeing the model from the constraints of graph structure, and enabling stronger semantic consistency modeling.

Moreover, this method is strictly nested within the unified aggregation paradigm of SDMP, inheriting its parameter-free, modular expressive advantages while achieving a good balance between execution efficiency and feature adaptability. As a key component of SDGNN, this strategy demonstrates excellent modeling performance across a variety of graph tasks, offering an effective path for the migration of structure-guided ideas to a lightweight feature space.

\subsubsection{Feature-Driven Pseudo Group Partitioning Strategy}
\label{subsec:theory}

In graphs with sparse structure, high noise, or cross-modal characteristics, topological relationships often deviate from the true semantics of nodes. This strategy and relies solely on feature similarity to construct neighborhood groups, enabling rapid aggregation with minimal computation. Its advantage lies in maintaining efficient execution without needing to analyze the structure, and it enhances semantic consistency in feature-driven scenarios. It is particularly suitable for tasks where structural signals are sparse, noisy, or unreliable, such as weakly connected networks and cross-modal data fusion.

To this end, we propose a completely feature-driven pseudo group partitioning strategy, aiming to eliminate reliance on structural information and construct neighbor groups in a more direct and lightweight manner, achieving fast and highly adaptive node aggregation. This strategy inherits the overall process of the dynamic partitioning framework proposed in the previous section. In contrast to structure-guided partitioning, it builds cluster centers solely from previous-layer features. The number of clusters can be specified independently or estimated from feature statistics, thus avoiding explicit reliance on structural cues. The entire process requires no iteration, no updates, and no additional parameters, fully reflecting the minimalist advantage of parameter-free aggregation.

Specifically, the strategy first fixes the number of clusters $K$ as the number of groups $S_v$ obtained from the structure-guided partitioning in Definition~3, but without utilizing its structural content:
\begin{equation}
	K = S_v.
\end{equation}
Then, for each neighbor node $u\in C_k$ in the initial cluster, we perform mean pooling on its previous-layer features $h_u^{(l-1)}$ to construct the initial center vector $c_k$:
\begin{equation}
	c_k = \frac{1}{\lvert C_k\rvert}\sum_{u\in C_k} h_u^{(l-1)}, \; k=1,2,\ldots,K.
\end{equation}
Based on these centers, we compute the Euclidean distance between all neighbor nodes and each center, assigning them to the nearest cluster, yielding a new non-overlapping set of neighbor clusters
\begin{equation}
	\mathcal{C}_v^{(l)}=\{C_1,C_2,\ldots,C_K\}, \; 
	u \mapsto \arg\min_{1\le k\le K}\big\|h_u^{(l-1)}-c_k\big\|_2 .
\end{equation}

Once the clusters are obtained, the aggregation step follows the SDMP framework in Eq.(9): each cluster’s mean feature is concatenated with the central node representation, and all cluster-level vectors are fused through cross-group max pooling and nonlinear activation to generate the final node embedding. This design preserves the parameter-free property while shifting the grouping criterion entirely to the feature space, ensuring consistency with the unified SDMP aggregation logic.

Although this mechanism may appear to be a special case of the structure-guided strategy, it fundamentally differs in modeling philosophy. Instead of relying on structural signals, it uses feature distributions as the sole basis for neighborhood partitioning. This design significantly improves adaptability in scenarios where structural information is weak, noisy, or unreliable—such as in weakly connected networks or cross-modal graphs. Moreover, since it avoids structural analysis and iterative center updates, the computational complexity is greatly reduced, enabling more efficient deployment on large-scale graph data.

Overall, this pseudo group partitioning strategy maintains the unified aggregation framework of SDMP in its implementation, but places greater emphasis on feature-driven modeling and minimal computation. It is a key lightweight supplement within the SDGNN architecture, providing a more flexible and efficient solution for graph modeling.

\subsection{Theoretical Analysis of Permutation Invariance of SDMP}
\label{subsec:sdmp-permutation}

To better characterize the generalization ability of SDMP, we further analyze its stability and robustness in terms of its expressive process. SDMP aggregation at each layer relies on three fundamental operations: mean pooling within groups, concatenation of node and group features, and cross-group max pooling. These operations inherently possess permutation invariance (see Theorem~1), meaning that no matter how the order of neighbor nodes changes within a group or across different groups, the final output representation $h_v^{(l)}$ remains stable. This property is particularly important for graph structures, where node neighbors are inherently unordered, and it significantly enhances the model's input robustness and generalization ability, especially in scenarios such as social networks and biological networks where structure is not labeled.

\medskip
\noindent\textbf{Theorem 1 (Permutation Invariance of SDMP).}
\label{thm:sdmp-perm1}
Let $G=(V,E)$ be a graph, with the neighbor set of node $v\in V$ denoted as $N(v)$, and the previous-layer feature representations as $\{h_u^{(l-1)} \mid u\in N(v)\}$. Let the group partitioning strategy be $\textsc{GroupPartition}(\cdot)$. If the group partitioning result $\mathcal{C}_v^{(l)}=\{C_{v,1},C_{v,2},\ldots,C_{v,S_v}\}$ does not change with the reordering of neighbors, then the representation $h_v^{(l)}$ computed by the structural-diversity message passing mechanism (SDMP) is invariant to the permutation of the neighbor inputs, i.e.,
\begin{equation}
	\forall\, \pi: N(v)\rightarrow N(v),\; \mathrm{SDMP}\!\left(N(v)\right)=\mathrm{SDMP}\!\left(\pi\!\left(N(v)\right)\right),
\end{equation}
where $\pi(\cdot)$ represents an arbitrary reordering of neighbors.

\noindent\textit{Proof.}
Since we assume that the group partitioning result remains unchanged after the reordering of neighbors, the subsequent computations are always based on the same group structure. The mean pooling operation within each group depends only on the feature values themselves, which is independent of the order of nodes within the group; the concatenation of node and group features is merely a vector-level operation, which does not include any order information; the max pooling operation across groups selects the maximum value dimension-wise, which is also independent of input order. Therefore, under the condition that group partitioning remains fixed, the final representation of SDMP is necessarily independent of the neighbor ordering, thus proving its permutation invariance. \hfill$\square$

\medskip
\noindent\textbf{Corollary 1 (Permutation Invariance of Structural Partitioning).}
When group partitioning is based on connected components of the neighbor-induced subgraph, the partitioning result is determined solely by the topological structure, and thus the output of SDMP under this strategy remains invariant to the neighbor ordering.

\noindent\textit{Proof.}
Connected components are entirely determined by the graph structure and are independent of input order. The subsequent operations, including mean pooling, concatenation, and max pooling, are all order-invariant, so the output remains unchanged. \hfill$\square$

\medskip
\noindent\textbf{Corollary 2 (Permutation Invariance of DBSCAN Clustering).}
When group partitioning employs DBSCAN clustering, under fixed parameters, the clustering result is uniquely determined by the semantic distance relationships between neighbors, and thus the output of SDMP under this strategy remains invariant to the neighbor ordering.

\noindent\textit{Proof.}
The core of DBSCAN lies in establishing density-based reachability relations using a distance matrix. This process is determined by the feature set and is independent of input order. The resulting cluster partitioning is deterministic. The subsequent operations of mean pooling, concatenation, and max pooling are also independent of order, so the final representation remains unchanged. \hfill$\square$

\medskip
\noindent\textbf{Corollary 3 (Permutation Invariance of Structure-Guided Dynamic Clustering).}
When group partitioning uses structure-guided dynamic clustering, the number of clusters is determined by the number of connected components in the first-order neighborhood, and the assignment of neighbors is based on their distance to the centers. Therefore, the result is independent of the neighbor ordering.

\noindent\textit{Proof.}
The number of clusters and the initial centers are uniquely determined by the graph structure and feature means. The assignment of neighbor nodes is determined by Euclidean distance, which is independent of order. The subsequent mean pooling and max pooling operations are also order-invariant, so the final representation remains unchanged. \hfill$\square$

\medskip
\noindent\textbf{Corollary 4 (Permutation Invariance of Feature-Driven Pseudo-Clustering).}
When group partitioning completely relies on feature-driven pseudo-clustering, the cluster centers are uniquely determined by the feature means, and the assignment of neighbor nodes is based on distance metrics. Therefore, the output of SDMP under this strategy is unaffected by the neighbor ordering.

\noindent\textit{Proof.}
The calculation of feature means is independent of the order of the set, and the clustering is uniquely determined by the distance relationships. The subsequent mean pooling and max pooling operations are also order-invariant, so the output satisfies permutation invariance. \hfill$\square$

\section{Structural Enhancement Mechanism Based on Normalized Propagation}
\label{sec:norm-prop}

In the previous chapters, we constructed three typical parameter-free aggregation paths based on local neighborhood group partitioning strategies, achieving effective coverage from structure-driven, feature-driven, to lightweight modeling approaches. However, node semantic representation is not only determined by local structure and feature interactions but is also influenced by its global position and cross-group relationships within the graph. Therefore, to further enhance the model’s ability to perceive long-range semantic dependencies and macro-structural features, we propose a normalized propagation-based structural enhancement mechanism as a global modeling supplement to the SDGNN framework.

This mechanism is inspired by the Simplified Graph Convolution (SGC) model \cite{ref13}, this mechanism achieves explicit encoding of multi-hop structural dependencies in the original graph by performing a one-time normalization and power propagation on the graph’s adjacency matrix, without introducing any trainable parameters. This results in a globally structure-aware node semantic representation.

Specifically, let the original graph be $G=(V,E)$, and the adjacency matrix be $A$. First, we add self-loops to each node, obtaining the enhanced adjacency matrix $\widetilde{A}=A+I$. Then, we perform symmetric normalization on $\widetilde{A}$ to obtain the propagation matrix:
\begin{equation}
	S = D^{-\frac{1}{2}}\,\widetilde{A}\,D^{-\frac{1}{2}},
\end{equation}
where $D$ is the degree matrix of $\widetilde{A}$. Next, we apply power $k$ (default $k=2$) to $S$ to integrate graph structural information from 1-hop to $k$-hop, resulting in the final propagation matrix:
\begin{equation}
	M = S^{k}.
\end{equation}
Using this propagation matrix, the initial features $X$ are transformed to obtain node representations that incorporate multi-hop structural information:
\begin{equation}
	X' = M X.
\end{equation}
In this process, multi-hop semantic integration is completed directly in the feature space, avoiding redundant computations associated with multi-layer stacking in traditional GCNs. The final $X'$ captures global context from long-range dependencies.

\medskip
\noindent\textbf{Theorem 2 (Multi-hop Structural Encoding of $\boldsymbol{S^{k}}$).}
For any positive integer $k$, the feature propagation result $X' = S^{k}X$, obtained from the propagation matrix $S^{k}$, explicitly encodes the structural information within at most $k$-hop neighborhoods without introducing any trainable parameters. Let the adjacency matrix of graph $G=(V,E)$ be $A$, and after self-loop enhancement and symmetric normalization, the propagation matrix is obtained as:
\begin{equation}
	S^{k} = \big(D^{-\frac{1}{2}}(A+I)D^{-\frac{1}{2}}\big)^{k}.
\end{equation}
Thus, for any positive integer $k$, the power propagation $X' = S^{k}X$ explicitly encodes the structural dependencies within at most $k$-hop neighborhoods without the need for any trainable parameters.

\noindent\textit{Proof.}
Based on the construction of the propagation matrix, we have
\begin{equation}
	S^{k} = \big(D^{-\frac{1}{2}}(A+I)D^{-\frac{1}{2}}\big)^{k}.
\end{equation}
Since $D^{-\frac{1}{2}}$ is a diagonal matrix, and diagonal matrices are commutative, we can extract it to both sides, simplifying it to:
\begin{equation}
	S^{k} = D^{-\frac{k}{2}}\,(A+I)^{k}\,D^{-\frac{k}{2}}.
\end{equation}
Using the binomial expansion formula,
\begin{equation}
	(A+I)^{k}=\sum_{i=0}^{k}\binom{k}{i}A^{i},
\end{equation}
we substitute to get:
\begin{equation}
	S^{k}=D^{-\frac{k}{2}}\left(\sum_{i=0}^{k}\binom{k}{i}A^{i}\right)D^{-\frac{k}{2}}.
\end{equation}
Here, the term $A^{i}$ describes the reachability relationship between nodes and their $i$-hop neighbors, indicating that this expansion shows that $S^{k}$ integrates all neighborhood structural information from 0-hop (self-loops) to $k$-hop. When $S^{k}$ acts on the node features $X$, the result $X' = S^{k}X$ explicitly encodes structural dependencies within up to $k$-hop neighborhoods, thus proving the multi-hop encoding capability of normalized propagation. \hfill$\square$

In practice, after completing normalized propagation, the enhanced features $X'$ are used as the node's initial features, and the structural partitioning strategy presented in Section~\ref{subsec:pf-group-partition} is applied: mean pooling is performed on each community, and the pooled results are concatenated with the center node. These are then passed through cross-group max pooling and activation functions to obtain the final node representation.

It is important to note that normalized propagation is only a one-time preprocessing step. The entire graph’s propagation matrix $S^{k}$ and the enhanced features $X'$ can be precomputed during model initialization and reused across all subsequent aggregation layers. This not only significantly reduces computational overhead but also avoids the potential information interference caused by repetitive modeling.

In summary, the normalized propagation-based structural enhancement mechanism achieves multi-hop semantic modeling with minimal computation. It significantly enhances the model’s ability to perceive global structures without introducing any training parameters. Through the propagation operation of $S^{k}$ (see Theorem~2), nodes can aggregate semantic information from their $k$-hop neighborhoods; the subsequent connected partitioning and local aggregation on the enhanced representations preserve the modeling ability of structural heterogeneity. This mechanism effectively integrates global propagation with local structural partitioning, balancing efficiency, interpretability, and expressiveness, making it an indispensable global modeling module in the SDGNN framework.

\section{Experiments}
\label{sec:experiments}

To comprehensively validate the effectiveness of the proposed SDGNN framework and its various group partitioning strategies across different tasks and scenarios, we designed a series of systematic experiments, covering node classification, interdisciplinary node recognition, and ablation analysis, among other aspects. All experiments are based on a unified model configuration and evaluation criteria, focusing on multiple datasets, tasks, and different training proportions. The goal is to deeply explore the differences in performance, computational efficiency, and applicability among the various group partitioning strategies. The following sections will first introduce the datasets used and the preprocessing methods, providing a foundation for the subsequent experimental results analysis.

\subsection{Data Preparation}
\label{sec:data-prep}

To comprehensively evaluate the adaptability and robustness of the proposed model under different graph structural complexities and feature distribution diversities, we selected eight representative public graph neural network datasets, covering multiple domains and task types, including academic networks (ACM \cite{ref28}, Cora \cite{ref29}, Citeseer \cite{ref29}, Pubmed \cite{ref29}, DBLP \cite{ref30}), movie networks (Film \cite{ref31}), course knowledge graphs (COCS \cite{ref32}), and geographic graphs (Texas \cite{ref31}).

These datasets exhibit significant differences in node scale, edge density, and the number of categories, making them ideal for validating the model's generalization ability across diverse graph data scenarios. All data are preprocessed into three components: adjacency matrix (Adjacency Matrix), node feature matrix (Feature Matrix), and node category labels (Label Vector), and are loaded in \texttt{.npy} format for subsequent training and evaluation. The basic statistical information of each dataset is shown in Table~\ref{tab:data-stats}.

\begin{table}
	\centering
	\caption{Basic statistics of different graph datasets.}
	\label{tab:data-stats}
	\footnotesize                     
	\setlength{\tabcolsep}{4pt}       
	\begin{tabular}{lccc}
		\hline
		\textbf{Dataset} & \textbf{Number of Nodes} & \textbf{Number of Edges} & \textbf{Number of Categories} \\
		\hline
		ACM     & 3025  & 13128 & 3 \\
		Citeseer& 3327  & 4614  & 6 \\
		COCS    & 18333 & 48535 & 15 \\
		Cora    & 2708  & 5278  & 7 \\
		DBLP    & 4057  & 3528  & 4 \\
		Film    & 7600  & 15009 & 5 \\
		Pubmed  & 19717 & 29845 & 3 \\
		Texas   & 183   & 162   & 5 \\
		\hline
	\end{tabular}
\end{table}

As can be seen from the table, the selected datasets differ significantly in terms of graph size, edge density, and category distribution, making them suitable for thoroughly testing the model's performance in various challenging scenarios such as low-resource, high-class-imbalance, and sparse graphs.

In the preprocessing of each dataset, we first constructed data objects based on PyTorch Geometric (PyG), including node feature matrices, edge indices, and label vectors. Then, we converted the graph structure to NetworkX format to pre-extract the connected component structure of the first-order neighborhood-induced subgraphs for each node. This information was cached as local community data for use in subsequent aggregation operations. This preprocessing step improves the computational efficiency of the feature aggregation phase and ensures that the model fully utilizes local graph constraints under the condition of no trainable parameters.

In the experimental design of this study, we adopted a two-layer structure as the standard model configuration for SDGNN and all its variants. That is, whether in classification experiments or interdisciplinary tasks, all models use a two-layer architecture. This choice is based on three considerations: First, the two-layer structure allows for effective integration of first-order neighborhood information and cross-cluster structural features while maintaining a lightweight computational process, thus balancing performance and efficiency; second, by using a uniform two-layer setup, we ensure comparability in the number of layers between SDGNN variants and compared parameterized models; and third, the two-layer structure has been widely validated as a reasonable choice in most public benchmark tasks, facilitating horizontal comparisons of experimental results. Therefore, this paper adopts the ``two-layer'' structure as the default configuration to support subsequent experiments and comparative analyses.

\subsection{Classification Experiments}
\label{sec:cls-exp}

To systematically evaluate the effectiveness and generalization ability of the four variants of SDGNN in node representation learning, we conducted large-scale node classification experiments and compared the results with various mainstream parameterized GNNs. The experiments cover multiple datasets and classification accuracy under different training ratios, and include analysis from two aspects: computational efficiency and node representation visualization, to comprehensively validate the advantages and limitations of SDGNN.

\subsubsection{Experimental Setup and Procedure}
\label{subsec:exp-setup}

To comprehensively evaluate the adaptability of different strategies under diverse network structures and feature distributions, we designed four variants of SDGNN and conducted node classification experiments. Among these, SDGNN-DBSCAN uses density-based clustering to adaptively identify locally dense groups in the neighborhood, preserving fine-grained differences in networks with significant structural heterogeneity. The SDGNN-DGS series (Dynamic Group Segmentation) is the focus of this paper. Its core idea is to use the structure-guided dynamic group partitioning mechanism to adaptively determine the number of clusters in local neighborhoods based on the initial structural branches, completing aggregation in the feature space. Specifically, DGS0 uses the initial partitioning results directly, emphasizing efficiency. DGS1 and {DGS2} perform one and two center update iterations, respectively, after the initial partitioning, to compare the trade-off between efficiency and accuracy with different iteration depths. It should be noted that DGS1 and DGS2 are not the core designs of the framework but are included for experimental comparison. In contrast, {SDGNN-FeatureOnly} relies entirely on the feature-driven, non-iterative pseudo-aggregation mechanism, discarding structural signals to emphasize semantic consistency and minimalist computation, making it suitable for scenarios with sparse structure or high noise. {SDGNN-SGCN}, on the other hand, introduces global enhancement based on normalized propagation (taking the second power) before local aggregation, explicitly encoding multi-hop neighborhood dependencies, and combines community partitioning to achieve complementary modeling of both global and local structures. All variants are fused using the Jumping Knowledge mechanism after multi-layer stacking to combine representations from different layers and are input to a lightweight two-layer MLP classifier for final prediction. A unified training and evaluation process ensures the comparability of different strategies under the same conditions and the reliability of the experimental conclusions.

To systematically assess the model’s generalization ability under varying levels of supervision, we set 9 different training set partition ratios for each dataset: $5\%$, $15\%$, $25\%$, $35\%$, $45\%$, $55\%$, $65\%$, $75\%$, and $85\%$. Let the total number of nodes in the graph be $n$. For each setting, we randomly select $\lfloor p \ast n \rfloor$ nodes from the entire $n$ nodes as the training set, where $p \in \{0.05,0.15,\ldots,0.85\}$. Then, from the remaining $n-\lfloor p \ast n \rfloor$ nodes, $10\%$ are partitioned into a validation set, and the rest are used as the test set. To reduce the impact of random sample partitioning on the experimental results, each configuration is run $5$ times, and the test accuracy is averaged. The entire procedure is uniformly executed across all datasets to ensure good comparability between results.

The entire classification process consists of two parts: feature aggregation and classifier training. First, the original node features are input into the corresponding SDGNN model, and the final node representations are generated through the aggregation mechanism. After obtaining the node representations, a two-layer perceptron (MLP) classifier is used for training, with the goal of minimizing the cross-entropy loss function. The classifier structure includes: one linear transformation layer (with an output dimension of $128$) combined with ReLU activation and Dropout (rate $=0.7$), followed by an output layer to generate category predictions. The Adam optimizer \cite{ref33} is used during training, with a learning rate of $0.01$, a weight decay coefficient of $1\times 10^{-4}$, and a maximum of $200$ training epochs. If the validation loss does not decrease for $30$ consecutive epochs, early stopping is applied to terminate training early. Finally, the trained classifier is used to predict node labels on the test set, and classification accuracy is used as the primary performance metric for the experiments.

\subsubsection{Classification Results and Efficiency Analysis}
\label{subsec:cls-results}

To comprehensively assess the differences in performance and efficiency among the various variants, we first conducted a systematic analysis of the classification accuracy for each method. We also introduced several mainstream graph neural network models as baseline comparisons, including: ComBSAGE, GCN, GAT, GraphSAGE, GIN, JK-NET, SAT, SGCN, and TANGNN. These models are based on publicly available implementations and were trained under the same feature inputs and data partitioning settings to ensure fairness in the comparison. The classification accuracy results of each model on different datasets and with various training ratios are shown in Tables~\ref{tab:acm}--\ref{tab:texas}, where the best results for each experiment are highlighted in bold.

\begin{table*}[ht]
	\centering
	\caption{Classification accuracy of different methods on the ACM dataset at varying training ratios (best per column in bold).}
	\label{tab:acm}
	\resizebox{\textwidth}{!}{%
		\begin{tabular}{lccccccccc}
			\hline
			\textbf{Method} & \textbf{5\%} & \textbf{15\%} & \textbf{25\%} & \textbf{35\%} & \textbf{45\%} & \textbf{55\%} & \textbf{65\%} & \textbf{75\%} & \textbf{85\%} \\
			\hline
			ComBSAGE & 0.6593 & 0.7220 & 0.7431 & 0.7581 & 0.7636 & 0.7738 & 0.7625 & 0.7785 & 0.7697 \\
			GAT & 0.7300 & 0.7824 & 0.7880 & 0.8017 & 0.8041 & 0.8104 & 0.8196 & 0.8233 & 0.8026 \\
			GCN & 0.7036 & 0.7527 & 0.7693 & 0.7807 & 0.7940 & 0.8025 & 0.8132 & 0.8079 & 0.7987 \\
			GIN & 0.6346 & 0.7126 & 0.7233 & 0.7246 & 0.7483 & 0.7536 & 0.7477 & 0.7433 & 0.7211 \\
			GraphSAGE & 0.5542 & 0.6199 & 0.6361 & 0.6434 & 0.6639 & 0.6606 & 0.6557 & 0.6708 & 0.6605 \\
			JK-NET & 0.7271 & 0.7724 & 0.7901 & 0.8040 & 0.8145 & 0.8096 & 0.8240 & 0.8334 & {0.8461} \\
			SAT & 0.5747 & 0.6358 & 0.6539 & 0.6638 & 0.6771 & 0.6813 & 0.7028 & 0.6826 & 0.6908 \\
			SGCN & 0.6966 & 0.7478 & 0.7683 & 0.7713 & 0.7692 & 0.7787 & 0.7799 & 0.7846 & 0.7987 \\
			TANGNN & 0.6708 & 0.8174 & 0.9037 & 0.9141 & 0.9156 & 0.9181 & 0.9165 & 0.9174 & 0.8750 \\
			SDGNN\_DBSCAN & 0.9023 & \textbf{0.9231} & \textbf{0.9276} & 0.9286 & 0.9275 & 0.9309 & 0.9334 & 0.9332 & 0.9289 \\
			SDGNN\_DGS0 & 0.8959 & 0.9159 & 0.9252 & 0.9267 & \textbf{0.9376} & 0.9343 & 0.9297 & 0.9327 & 0.9263 \\
			SDGNN\_DGS1 & 0.9026 & 0.9175 & 0.9236 & 0.9247 & 0.9286 & 0.9262 & 0.9276 & 0.9284 & 0.9368 \\
			SDGNN\_DGS2 & 0.8963 & 0.9209 & 0.9248 & \textbf{0.9295} & 0.9302 & 0.9300 & 0.9318 & \textbf{0.9367} & 0.9368 \\
			SDGNN\_FeatureOnly & \textbf{0.9047} & 0.9222 & 0.9247 & 0.9262 & 0.9294 & \textbf{0.9360} & \textbf{0.9337} & 0.9363 & \textbf{0.9421} \\
			SDGNN\_SGCN & 0.8867 & 0.9036 & 0.9103 & 0.9109 & 0.9191 & 0.9181 & 0.9226 & 0.9213 & 0.9263 \\
			\hline
	\end{tabular}}%
\end{table*}

\begin{table*}[ht]
	\centering
	\caption{Classification accuracy of different methods on the Citeseer dataset at varying training ratios (best per column in bold).}
	\label{tab:citeseer}
	\resizebox{\textwidth}{!}{%
		\begin{tabular}{lccccccccc}
			\hline
			{Method} & \textbf{5\%} & \textbf{15\%} & \textbf{25\%} & \textbf{35\%} & \textbf{45\%} & \textbf{55\%} & \textbf{65\%} & \textbf{75\%} & \textbf{85\%} \\
			\hline
			ComBSAGE & 0.4253 & 0.5156 & 0.5396 & 0.5601 & 0.5750 & 0.5741 & 0.5906 & 0.5976 & 0.5881 \\
			GAT & 0.4757 & 0.5416 & 0.5716 & 0.5993 & 0.6262 & 0.6415 & 0.6363 & 0.6404 & 0.6333 \\
			GCN & 0.5041 & 0.5732 & 0.6056 & 0.6321 & 0.6417 & 0.6436 & 0.6744 & 0.6856 & 0.6702 \\
			GIN & 0.4232 & 0.5004 & 0.5178 & 0.5337 & 0.5354 & 0.5417 & 0.5551 & 0.5496 & 0.5714 \\
			GraphSAGE & 0.3668 & 0.4375 & 0.4608 & 0.4780 & 0.4858 & 0.4894 & 0.5030 & 0.4940 & 0.5119 \\
			JK-NET & 0.4476 & 0.5325 & 0.5622 & 0.5827 & 0.6016 & 0.6045 & 0.6170 & 0.6124 & 0.6536 \\
			SAT & 0.3465 & 0.4079 & 0.4361 & 0.4594 & 0.4665 & 0.4818 & 0.4814 & 0.4788 & 0.5083 \\
			SGCN & 0.3540 & 0.4712 & 0.5036 & 0.5317 & 0.5314 & 0.5280 & 0.5248 & 0.5564 & 0.5274 \\
			TANGNN & 0.5648 & 0.6727 & 0.7009 & 0.7032 & 0.7174 & 0.7300 & 0.7359 & 0.7304 & 0.7488 \\
			SDGNN\_DBSCAN & 0.6581 & 0.7046 & 0.7174 & 0.7288 & \textbf{0.7406} & 0.7446 & 0.7561 & 0.7616 & 0.7667 \\
			SDGNN\_DGS0 & 0.6568 & 0.7058 & 0.7222 & 0.7302 & 0.7350 & 0.7436 & 0.7465 & \textbf{0.7628} & 0.7607 \\
			SDGNN\_DGS1 & 0.6524 & 0.7071 & 0.7231 & 0.7317 & 0.7366 & 0.7496 & 0.7381 & 0.7360 & 0.7548 \\
			SDGNN\_DGS2 & 0.6565 & 0.6985 & 0.7126 & 0.7275 & 0.7382 & 0.7441 & 0.7484 & 0.7468 & \textbf{0.7845} \\
			SDGNN\_FeatureOnly & \textbf{0.6777} & \textbf{0.7168} & \textbf{0.7218} & 0.7304 & 0.7405 & \textbf{0.7530} & 0.7467 & 0.7624 & 0.7810 \\
			SDGNN\_SGCN & 0.6713 & 0.6987 & 0.7135 & \textbf{0.7324} & 0.7283 & 0.7410 & \textbf{0.7412} & 0.7428 & 0.7786 \\
			\hline
	\end{tabular}}%
\end{table*}

\begin{table*}[ht]
	\centering
	\caption{Classification accuracy of different methods on the COCS dataset at varying training ratios (best per column in bold).}
	\label{tab:cocs}
	\resizebox{\textwidth}{!}{%
		\begin{tabular}{lccccccccc}
			\hline
			\textbf{Method} & \textbf{5\%} & \textbf{15\%} & \textbf{25\%} & \textbf{35\%} & \textbf{45\%} & \textbf{55\%} & \textbf{65\%} & \textbf{75\%} & \textbf{85\%} \\
			\hline
			ComBSAGE & 0.6071 & 0.6466 & 0.6548 & 0.6624 & 0.6613 & 0.6709 & 0.6661 & 0.6667 & 0.6872 \\
			GAT & 0.5863 & 0.6127 & 0.6193 & 0.6207 & 0.6217 & 0.6220 & 0.6277 & 0.6329 & 0.6279 \\
			GCN & 0.5764 & 0.6063 & 0.6089 & 0.6149 & 0.6172 & 0.6192 & 0.6161 & 0.6198 & 0.6240 \\
			GIN & 0.5258 & 0.5605 & 0.5658 & 0.5711 & 0.5708 & 0.5791 & 0.5784 & 0.5844 & 0.5841 \\
			GraphSAGE & 0.5507 & 0.5881 & 0.5941 & 0.5982 & 0.6057 & 0.6061 & 0.6059 & 0.6166 & 0.6268 \\
			JK-NET & 0.6284 & 0.6663 & 0.6697 & 0.6789 & 0.6847 & 0.6856 & 0.6858 & 0.6904 & 0.6914 \\
			SAT & 0.5343 & 0.5661 & 0.5792 & 0.5832 & 0.5849 & 0.5936 & 0.5880 & 0.5935 & 0.5893 \\
			SGCN & 0.3999 & 0.4555 & 0.4721 & 0.4747 & 0.4940 & 0.4918 & 0.4880 & 0.4971 & 0.4892 \\
			TANGNN & 0.8338 & 0.8773 & 0.9028 & 0.9094 & 0.9134 & 0.9216 & 0.9162 & 0.9250 & 0.9138 \\
			SDGNN\_DBSCAN & \textbf{0.8797} & \textbf{0.9145} & \textbf{0.9254} & 0.9287 & \textbf{0.9342} & \textbf{0.9366} & \textbf{0.9389} & \textbf{0.9416} & 0.9446 \\
			SDGNN\_DGS0 & 0.8743 & 0.9075 & 0.9181 & 0.9253 & 0.9291 & 0.9292 & 0.9360 & 0.9370 & 0.9348 \\
			SDGNN\_DGS1 & 0.8746 & 0.9053 & 0.9183 & 0.9252 & 0.9292 & 0.9306 & 0.9349 & 0.9373 & \textbf{0.9477} \\
			SDGNN\_DGS2 & 0.8767 & 0.9077 & 0.9194 & 0.9232 & 0.9306 & 0.9295 & 0.9349 & 0.9357 & 0.9350 \\
			SDGNN\_FeatureOnly & 0.8788 & 0.9083 & 0.9201 & \textbf{0.9248} & 0.9313 & 0.9311 & 0.9363 & 0.9397 & 0.9444 \\
			SDGNN\_SGCN & 0.7609 & 0.7824 & 0.7882 & 0.7926 & 0.7963 & 0.7951 & 0.7991 & 0.8114 & 0.7965 \\
			\hline
	\end{tabular}}%
\end{table*}

\begin{table*}[ht]
	\centering
	\caption{Classification accuracy of different methods on the Cora dataset at varying training ratios (best per column in bold).}
	\label{tab:cora}
	\resizebox{\textwidth}{!}{%
		\begin{tabular}{lccccccccc}
			\hline
			\textbf{Method} & \textbf{5\%} & \textbf{15\%} & \textbf{25\%} & \textbf{35\%} & \textbf{45\%} & \textbf{55\%} & \textbf{65\%} & \textbf{75\%} & \textbf{85\%} \\
			\hline
			ComBSAGE & 0.4822 & 0.5932 & 0.6301 & 0.6601 & 0.6730 & 0.6896 & 0.6965 & 0.6973 & 0.7124 \\
			GAT & 0.6677 & 0.7495 & 0.7706 & 0.7893 & 0.8046 & 0.8126 & 0.8032 & 0.8152 & 0.8307 \\
			GCN & 0.6272 & 0.7173 & 0.7456 & 0.7641 & 0.7798 & 0.7954 & 0.7985 & 0.8044 & 0.7854 \\
			GIN & 0.5571 & 0.6268 & 0.6577 & 0.6732 & 0.6852 & 0.6929 & 0.6988 & 0.6968 & 0.6701 \\
			GraphSAGE & 0.4390 & 0.5008 & 0.5433 & 0.5616 & 0.5782 & 0.5842 & 0.6015 & 0.5769 & 0.6219 \\
			JK-NET & 0.5546 & 0.6632 & 0.6880 & 0.7068 & 0.7331 & 0.7305 & 0.7434 & 0.7273 & 0.7635 \\
			SAT & 0.4173 & 0.4928 & 0.5356 & 0.5579 & 0.5584 & 0.5619 & 0.5764 & 0.5862 & 0.5839 \\
			SGCN & 0.4333 & 0.5335 & 0.5663 & 0.6091 & 0.5989 & 0.6192 & 0.6339 & 0.6339 & 0.6365 \\
			TANGNN & 0.6987 & 0.7857 & 0.8294 & 0.8393 & 0.8436 & 0.8573 & 0.8652 & 0.8629 & 0.8745 \\
			SDGNN\_DBSCAN & 0.7772 & 0.8311 & 0.8420 & 0.8538 & 0.8587 & 0.8655 & 0.8764 & 0.8713 & 0.8555 \\
			SDGNN\_DGS0 & 0.7515 & 0.8194 & 0.8343 & 0.8474 & 0.8539 & 0.8590 & 0.8643 & 0.8624 & 0.8526 \\
			SDGNN\_DGS1 & 0.7434 & 0.8193 & 0.8374 & 0.8445 & 0.8487 & 0.8563 & 0.8566 & 0.8673 & 0.8380 \\
			SDGNN\_DGS2 & 0.7567 & 0.8111 & 0.8342 & 0.8424 & 0.8502 & 0.8550 & 0.8640 & 0.8624 & 0.8569 \\
			SDGNN\_FeatureOnly & 0.7508 & 0.8169 & 0.8408 & 0.8479 & 0.8536 & 0.8679 & 0.8631 & 0.8747 & \textbf{0.8759} \\
			SDGNN\_SGCN & \textbf{0.8079} & \textbf{0.8409} & \textbf{0.8507} & \textbf{0.8629} & \textbf{0.8590} & \textbf{0.8717} & \textbf{0.8776} & \textbf{0.8826} & 0.8701 \\
			\hline
	\end{tabular}}%
\end{table*}

\begin{table*}[ht]
	\centering
	\caption{Classification accuracy of different methods on the DBLP dataset at varying training ratios (best per column in bold).}
	\label{tab:dblp}
	\resizebox{\textwidth}{!}{%
		\begin{tabular}{lccccccccc}
			\hline
			\textbf{Method} & \textbf{5\%} & \textbf{15\%} & \textbf{25\%} & \textbf{35\%} & \textbf{45\%} & \textbf{55\%} & \textbf{65\%} & \textbf{75\%} & \textbf{85\%} \\
			\hline
			ComBSAGE & 0.5796 & 0.6399 & 0.6679 & 0.6780 & 0.6960 & 0.6922 & 0.6999 & 0.7200 & 0.7049 \\
			GAT & 0.6049 & 0.6555 & 0.6798 & 0.6919 & 0.7147 & 0.7239 & 0.7245 & 0.7302 & 0.7255 \\
			GCN & 0.6248 & 0.6706 & 0.6939 & 0.7023 & 0.7151 & 0.7282 & 0.7452 & 0.7456 & 0.7578 \\
			GIN & 0.5094 & 0.5566 & 0.5879 & 0.5931 & 0.6053 & 0.6083 & 0.6158 & 0.6026 & 0.6000 \\
			GraphSAGE & 0.5523 & 0.6025 & 0.6327 & 0.6411 & 0.6477 & 0.6474 & 0.6709 & 0.6715 & 0.6931 \\
			JK-NET & 0.6267 & 0.6802 & 0.7062 & 0.7115 & 0.7205 & 0.7346 & 0.7391 & 0.7436 & 0.7490 \\
			SAT & 0.5117 & 0.5743 & 0.5916 & 0.6089 & 0.6253 & 0.6276 & 0.6343 & 0.6472 & 0.6206 \\
			SGCN & 0.5897 & 0.6544 & 0.6703 & 0.6834 & 0.6925 & 0.7016 & 0.6857 & 0.7007 & 0.6941 \\
			TANGNN & 0.7077 & 0.7777 & 0.7889 & 0.8006 & 0.7317 & 0.8218 & 0.8217 & 0.8220 & 0.8216 \\
			SDGNN\_DBSCAN & 0.7602 & 0.7997 & 0.8149 & 0.8203 & \textbf{0.8252} & \textbf{0.8336} & 0.8355 & 0.8252 & \textbf{0.8608} \\
			SDGNN\_DGS0 & 0.7671 & 0.7902 & 0.8071 & 0.8193 & 0.8261 & 0.8334 & 0.8355 & \textbf{0.8531} & 0.8461 \\
			SDGNN\_DGS1 & 0.7694 & 0.7984 & 0.8125 & 0.8194 & 0.8248 & 0.8262 & 0.8319 & 0.8357 & 0.8216 \\
			SDGNN\_DGS2 & 0.7679 & 0.7992 & 0.8158 & 0.8191 & 0.8157 & 0.8258 & \textbf{0.8370} & 0.8315 & 0.8245 \\
			SDGNN\_FeatureOnly & \textbf{0.7744} & \textbf{0.8028} & \textbf{0.8152} & \textbf{0.8236} & 0.8250 & 0.8270 & 0.8394 & 0.8393 & 0.8539 \\
			SDGNN\_SGCN & 0.7481 & 0.7825 & 0.7971 & 0.8060 & 0.8164 & 0.8153 & 0.8191 & 0.8246 & 0.8216 \\
			\hline
	\end{tabular}}%
\end{table*}

\begin{table*}[ht]
	\centering
	\caption{Classification accuracy of different methods on the Film dataset at varying training ratios (best per column in bold).}
	\label{tab:film}
	\resizebox{\textwidth}{!}{%
		\begin{tabular}{lccccccccc}
			\hline
			\textbf{Method} & \textbf{5\%} & \textbf{15\%} & \textbf{25\%} & \textbf{35\%} & \textbf{45\%} & \textbf{55\%} & \textbf{65\%} & \textbf{75\%} & \textbf{85\%} \\
			\hline
			ComBSAGE & 0.2495 & 0.2538 & 0.2563 & 0.2649 & 0.2612 & 0.2611 & 0.2745 & 0.2588 & 0.2716 \\
			GAT & 0.2699 & 0.2838 & 0.2853 & 0.2856 & 0.2902 & 0.2872 & 0.2901 & 0.2896 & 0.2700 \\
			GCN & 0.2645 & 0.2682 & 0.2694 & 0.2728 & 0.2722 & 0.2757 & 0.2762 & 0.2830 & \textbf{0.2926} \\
			GIN & 0.2514 & 0.2496 & 0.2574 & 0.2521 & 0.2590 & 0.2528 & 0.2521 & 0.2604 & 0.2553 \\
			GraphSAGE & 0.2511 & 0.2539 & 0.2630 & {0.2949} & 0.2762 & {0.3000} & {0.3068} & {0.3089} & 0.2811 \\
			JK-NET & 0.2628 & 0.2759 & 0.2810 & 0.2795 & 0.2850 & 0.2746 & 0.2782 & 0.2888 & 0.2853 \\
			SAT & 0.2430 & 0.2547 & 0.2593 & 0.2559 & 0.2740 & 0.2888 & 0.2896 & 0.2875 & 0.2689 \\
			SGCN & 0.2487 & 0.2555 & 0.2544 & 0.2566 & 0.2527 & 0.2619 & 0.2559 & 0.2609 & 0.2326 \\
			TANGNN & 0.2642 & 0.2593 & {0.2892} & 0.2748 & 0.2571 & 0.2802 & 0.2880 & 0.2770 & 0.2763 \\
			SDGNN\_DBSCAN & \textbf{0.2737} & \textbf{0.2956} & \textbf{0.3094} & \textbf{0.3148} & \textbf{0.3177} & \textbf{0.3315} & \textbf{0.3437} & \textbf{0.3398} & \textbf{0.3521} \\
			SDGNN\_DGS0 & 0.2614 & 0.2878 & 0.2950 & 0.3029 & 0.3077 & 0.3177 & 0.3188 & 0.3368 & 0.3363 \\
			SDGNN\_DGS1 & 0.2659 & 0.2867 & 0.2992 & 0.3036 & 0.3051 & 0.3210 & 0.3249 & 0.3272 & 0.3153 \\
			SDGNN\_DGS2 & 0.2530 & \textbf{0.2956} & 0.2974 & 0.3108 & 0.3137 & 0.3194 & 0.3227 & 0.3226 & 0.3384 \\
			SDGNN\_FeatureOnly & 0.2674 & 0.2930 & 0.3015 & 0.3139 & 0.3190 & 0.3189 & 0.3278 & 0.3372 & 0.3416 \\
			SDGNN\_SGCN & 0.2488 & 0.2664 & 0.2671 & 0.2702 & 0.2678 & 0.2691 & 0.2832 & 0.2818 & 0.2874 \\
			\hline
	\end{tabular}}%
\end{table*}

\begin{table*}[ht]
	\centering
	\caption{Classification accuracy of different methods on the Pubmed dataset at varying training ratios (best per column in bold).}
	\label{tab:pubmed}
	\resizebox{\textwidth}{!}{%
		\begin{tabular}{lccccccccc}
			\hline
			\textbf{Method} & \textbf{5\%} & \textbf{15\%} & \textbf{25\%} & \textbf{35\%} & \textbf{45\%} & \textbf{55\%} & \textbf{65\%} & \textbf{75\%} & \textbf{85\%} \\
			\hline
			ComBSAGE & 0.3949 & 0.3987 & 0.3987 & 0.3968 & 0.3956 & 0.3927 & 0.4017 & 0.4032 & 0.3895 \\
			GAT & 0.5888 & 0.6009 & 0.6057 & 0.6056 & 0.6052 & 0.6109 & 0.6097 & 0.6124 & 0.6041 \\
			GCN & 0.5754 & 0.5888 & 0.5946 & 0.5970 & 0.5977 & 0.5995 & 0.5983 & 0.6056 & 0.5966 \\
			GIN & 0.4769 & 0.5031 & 0.5052 & 0.5090 & 0.5073 & 0.5111 & 0.5098 & 0.5089 & 0.5319 \\
			GraphSAGE & 0.3932 & 0.4235 & 0.4636 & 0.4560 & 0.4625 & 0.4614 & 0.4842 & 0.4951 & 0.4721 \\
			JK-NET & 0.4363 & 0.5746 & 0.5938 & 0.5972 & 0.6086 & 0.6190 & 0.6112 & 0.6086 & 0.6079 \\
			SAT & 0.3930 & 0.4427 & 0.5023 & 0.5094 & 0.5306 & 0.5287 & 0.5312 & 0.5461 & 0.5627 \\
			SGCN & 0.3991 & 0.3981 & 0.3986 & 0.3928 & 0.3936 & 0.3996 & 0.4036 & 0.3958 & 0.3939 \\
			TANGNN & 0.3938 & 0.3947 & 0.3987 & 0.3982 & 0.4868 & 0.3937 & 0.4869 & 0.3940 & 0.5741 \\
			SDGNN\_DBSCAN & \textbf{0.7925} & \textbf{0.8277} & \textbf{0.8462} & \textbf{0.8563} & \textbf{0.8610} & \textbf{0.8683} & \textbf{0.8714} & \textbf{0.8764} & 0.8699 \\
			SDGNN\_DGS0 & 0.7781 & 0.8198 & 0.8369 & 0.8459 & 0.8557 & 0.8594 & 0.8629 & 0.8658 & 0.8659 \\
			SDGNN\_DGS1 & 0.7789 & 0.8176 & 0.8363 & 0.8466 & 0.8573 & 0.8595 & 0.8639 & 0.8698 & 0.8687 \\
			SDGNN\_DGS2 & 0.7768 & 0.8173 & 0.8361 & 0.8466 & 0.8581 & 0.8618 & 0.8665 & 0.8674 & 0.8582 \\
			SDGNN\_FeatureOnly & 0.7814 & 0.8201 & 0.8382 & 0.8502 & 0.8573 & 0.8623 & 0.8654 & 0.8727 & \textbf{0.8806} \\
			SDGNN\_SGCN & 0.6792 & 0.6997 & 0.7073 & 0.7111 & 0.7145 & 0.7125 & 0.7167 & 0.7222 & 0.7147 \\
			\hline
	\end{tabular}}%
\end{table*}

\begin{table*}[ht]
	\centering
	\caption{Classification accuracy of different methods on the Texas dataset at varying training ratios (best per column in bold).}
	\label{tab:texas}
	\resizebox{\textwidth}{!}{%
		\begin{tabular}{lccccccccc}
			\hline
			\textbf{Method} & \textbf{5\%} & \textbf{15\%} & \textbf{25\%} & \textbf{35\%} & \textbf{45\%} & \textbf{55\%} & \textbf{65\%} & \textbf{75\%} & \textbf{85\%} \\
			\hline
			ComBSAGE & 0.4218 & 0.5536 & 0.5567 & 0.5564 & 0.5518 & 0.5446 & 0.5489 & 0.5143 & 0.5400 \\
			GAT & 0.4782 & 0.5348 & 0.5567 & 0.5366 & 0.5325 & 0.5262 & 0.5319 & 0.5571 & 0.6000 \\
			GCN & 0.5474 & 0.5435 & 0.5617 & 0.5584 & 0.5542 & 0.5692 & 0.5234 & 0.5786 & 0.5400 \\
			GIN & 0.5385 & 0.5652 & 0.5567 & 0.5426 & 0.5518 & 0.5508 & 0.5447 & 0.5643 & 0.5400 \\
			GraphSAGE & 0.5474 & 0.5464 & 0.5550 & 0.5604 & 0.5855 & 0.5477 & 0.5574 & 0.5429 & 0.6600 \\
			JK-NET & 0.5590 & 0.5348 & 0.5300 & 0.5683 & 0.5566 & 0.5692 & 0.6000 & 0.5357 & 0.6400 \\
			SAT & 0.5885 & 0.5812 & 0.5867 & 0.5762 & 0.5928 & 0.6215 & 0.6043 & 0.5357 & 0.5200 \\
			SGCN & 0.3987 & 0.5507 & 0.5683 & 0.5762 & 0.5759 & 0.5323 & 0.5362 & 0.5357 & 0.4800 \\
			TANGNN & 0.5269 & 0.4609 & 0.5867 & 0.5723 & 0.6651 & 0.6492 & 0.5617 & 0.6857 & 0.7600 \\
			SDGNN\_DBSCAN & 0.4667 & 0.6188 & 0.7217 & 0.7446 & 0.7687 & 0.7538 & 0.8000 & 0.8000 & 0.8400 \\
			SDGNN\_DGS0 & 0.5654 & 0.6551 & 0.7350 & 0.7564 & 0.7687 & 0.7969 & 0.7957 & 0.8286 & \textbf{0.9400} \\
			SDGNN\_DGS1 & 0.5654 & 0.6681 & 0.6867 & \textbf{0.7743} & 0.7687 & \textbf{0.8154} & 0.7532 & \textbf{0.8643} & 0.8400 \\
			SDGNN\_DGS2 & 0.5872 & 0.6391 & 0.7083 & 0.7446 & 0.7542 & 0.8062 & \textbf{0.8426} & 0.7357 & 0.8000 \\
			SDGNN\_FeatureOnly & \textbf{0.6218} & \textbf{0.6899} & \textbf{0.7600} & 0.7723 & \textbf{0.7952} & 0.7969 & 0.8043 & 0.8429 & 0.9000 \\
			SDGNN\_SGCN & 0.5462 & 0.5638 & 0.5400 & 0.5703 & 0.5542 & 0.5262 & 0.5957 & 0.6071 & 0.5200 \\
			\hline
	\end{tabular}}%
\end{table*}

Overall, the results show that the SDGNN series significantly outperforms traditional parameterized GNNs on most datasets, particularly in graphs with complex structures or significant category differences (e.g., Pubmed, COCS, and Texas). For example, on the Pubmed dataset, {SDGNN-FeatureOnly} achieves the highest accuracy of $0.8806$ at the $85\%$ training ratio (Table~\ref{tab:pubmed}), far surpassing the next best-performing TANGNN ($0.5741$). On the Texas dataset, {SDGNN-DGS0} achieves an accuracy of $0.94$ at the $85\%$ training ratio (Table~\ref{tab:texas}), setting a new best result among all comparison methods.

Different variants exhibit unique performance characteristics. {SDGNN-DBSCAN} performs overall best on networks with significant neighborhood density differences, such as Citeseer, COCS, and Pubmed (Tables~\ref{tab:citeseer}--\ref{tab:pubmed}), demonstrating that the density-based clustering mechanism is better suited to structural heterogeneity. {SDGNN-DGS0}, as the original design, consistently approaches or even exceeds the performance of DBSCAN on several datasets and achieves the highest accuracy on Texas. Its counterpart versions, {DGS1/2}, show slight improvements under certain training ratios but generally follow a similar trend to DGS0. {SDGNN-FeatureOnly} excels in scenarios dominated by high-dimensional features, achieving leading results on datasets like Pubmed and Texas, highlighting the semantic consistency driven by features. {SDGNN-SGCN}, while slightly less accurate, remains competitive on datasets such as ACM and Cora (Tables~\ref{tab:acm} and \ref{tab:cora}), showcasing the effectiveness of global enhancement in its modeling.

In comparison, parameterized models such as GAT and JK-NET achieve competitive results on small-scale or class-balanced graphs (e.g., ACM and Cora), but their performance drops significantly on graphs with highly imbalanced class distributions (e.g., Texas) or sparse node features (e.g., Pubmed). Thanks to the parameter-free two-stage aggregation mechanism, the SDGNN series can adapt to the diversity of node features and structures, effectively mitigating the risk of overfitting, and maintaining strong generalization ability in low-supervision scenarios. For example, on the Citeseer dataset, with only $5\%$ of the training data, {SDGNN-DBSCAN} achieves an accuracy of $0.6581$ (Table~\ref{tab:citeseer}), far surpassing GCN ($0.5041$) and GAT ($0.4757$).

In conclusion, the SDGNN series, through structure-guided clustering partitioning and cross-cluster feature fusion, balances accuracy and efficiency in most node classification tasks. Among them, {DBSCAN} is suitable for networks with strong structural heterogeneity, {DGS0} is ideal for most general tasks, {FeatureOnly} highlights semantic consistency, and {SGCN} offers extremely high efficiency for large-scale graphs. While still slightly slower than some parameterized GNNs, SDGNN achieves superior performance and cross-scenario adaptability with lower model complexity, demonstrating its potential for application in non-parameterized graph learning.

To further compare the efficiency of each strategy, we calculated the average feature aggregation time (in seconds) across different datasets, with the results shown in Table~\ref{tab:agg-time}.

\begin{table*}[ht]
	\centering
	\caption{Average feature aggregation time (in seconds) for different group partitioning strategies.}
	\label{tab:agg-time}
	\resizebox{\textwidth}{!}{%
		\begin{tabular}{lcccccc}
			\hline
			\textbf{Dataset} & \textbf{SDGNN\_DBSCAN} & \textbf{SDGNN\_DGS0} & \textbf{SDGNN\_DGS1} & \textbf{SDGNN\_DGS2} & \textbf{SDGNN\_FeatureOnly} & \textbf{SDGNN\_SGCN} \\
			\hline
			ACM      & 13.5316 & 3.4021 & 9.5770  & 10.0507 & \textbf{2.8635} & 2.9540 \\
			Citeseer & 10.8768 & 1.4479 & 12.8712 & 11.7642 & 2.0856 & \textbf{0.5486} \\
			COCS     & 326.8331 & 6.8039 & 84.0595 & 83.1415 & 11.8313 & \textbf{7.4646} \\
			Cora     & 8.6889  & 1.2312 & 9.6899  & 9.6971  & 1.8173 & \textbf{0.8260} \\
			DBLP     & 3.1460  & 1.0066 & 8.9206  & 8.8398  & 1.7569 & \textbf{0.5560} \\
			Film     & 24.5039 & 4.7156 & 31.5340 & 31.8735 & 6.7920 & \textbf{4.6929} \\
			Pubmed   & 38.4632 & 7.9787 & 69.6982 & 70.0739 & 12.2211 & \textbf{6.2513} \\
			Texas    & 0.3705  & 0.0840 & 0.6393  & 0.6681  & 0.1190 & \textbf{0.0230} \\
			\hline
	\end{tabular}}%
\end{table*}

The results show that DBSCAN aggregation takes significant time on large-scale graphs (e.g., COCS and Pubmed), mainly due to the need for multiple neighborhood density estimations and global distance calculations. This complexity increases further in high-dimensional feature spaces, limiting its scalability. In contrast, non-iterative DGS (SDGNN-DGS0) significantly reduces aggregation time across all datasets, especially on large-scale graphs like COCS and Pubmed, making it the core solution that balances efficiency and performance. DGS1 and DGS2 (which perform one and two center update iterations, respectively) can improve accuracy in some cases, but since each iteration requires recalculating the distances from nodes to cluster centers, their time complexity increases linearly or even super-linearly with the number of iterations, ultimately slowing down the overall training process. Therefore, only the non-iterative version (DGS0) is retained as the default in the final design.

Additionally, SDGNN-FeatureOnly is slightly less efficient than DGS0 but avoids potential biases introduced by structural partitioning, offering an advantage in semantic consistency. SDGNN-SGCN, by leveraging multi-hop normalized adjacency matrix propagation, effectively reduces the computational burden of local clustering. Its efficiency even surpasses that of DGS0 on large-scale graphs such as COCS and Pubmed, making it the most scalable aggregation strategy.

It should be noted that while the SDGNN series outperforms most parameterized GNNs in classification performance, it does not have an absolute advantage in terms of runtime. Traditional methods (e.g., GCN, GAT), using standard convolution or attention mechanisms, skip the community detection and clustering steps, allowing for rapid training and inference on GPUs. In contrast, although SDGNN does not require parameter training, the steps of community detection, clustering, and cross-cluster integration inevitably introduce additional computational overhead, especially on large-scale graphs.

\subsubsection{Node Representation Visualization Analysis}
\label{subsec:vis}

To further evaluate the performance of different methods in the node representation learning task and visually demonstrate their impact on intra-class compactness and inter-class separability, we performed node representation visualization on the Cora dataset. We compared the four SDGNN variants with eight mainstream baseline models (ComBSAGE, GCN, GAT, GraphSAGE, GIN, JK-Net, SAT, SGCN, TANGNN). The specific procedure is as follows: First, we obtained the node embeddings after completing the classification experiments for each model, and then reduced them to a two-dimensional plane using t-SNE \cite{ref34}. Next, we color-coded the nodes according to their true class labels to visually present the distribution characteristics of different classes of nodes in the embedding space.

\begin{figure}[p]  
	\centering

	\newcommand{\panelimg}[1]{\includegraphics[width=\linewidth,height=0.21\textheight,keepaspectratio]{#1}\vspace{2pt}}
	
	\begin{minipage}[b]{0.48\linewidth}\centering
		\panelimg{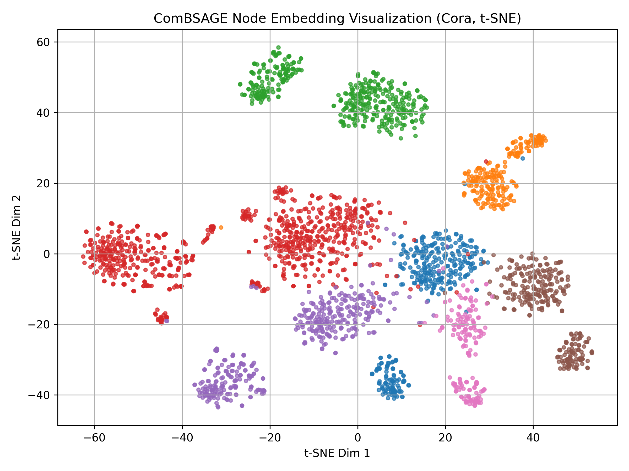}\\
		\scriptsize (a) ComBSAGE
	\end{minipage}\hfill
	\begin{minipage}[b]{0.48\linewidth}\centering
		\panelimg{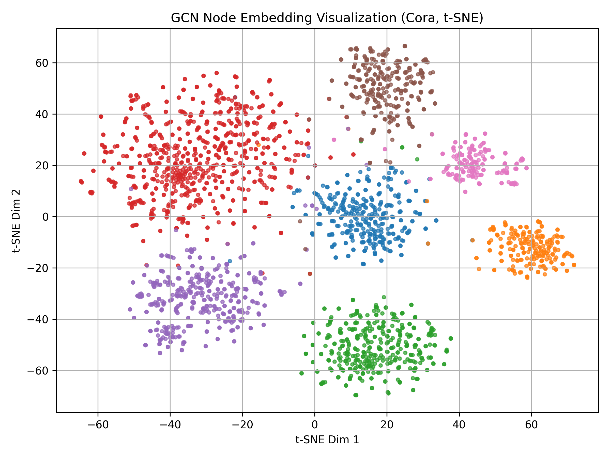}\\
		\scriptsize (b) GCN
	\end{minipage}
	
	\vspace{6pt}
	\begin{minipage}[b]{0.48\linewidth}\centering
		\panelimg{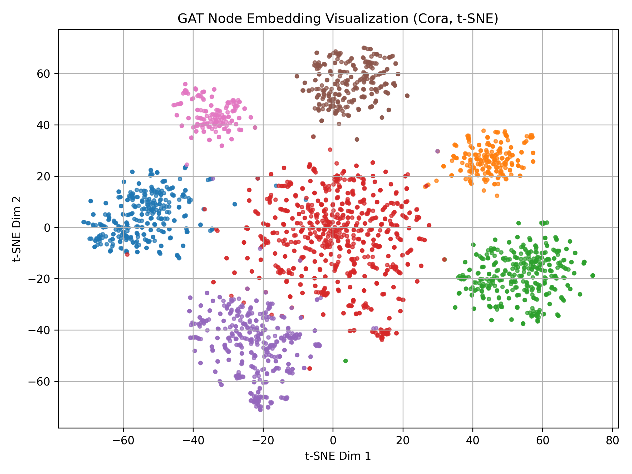}\\
		\scriptsize (c) GAT
	\end{minipage}\hfill
	\begin{minipage}[b]{0.48\linewidth}\centering
		\panelimg{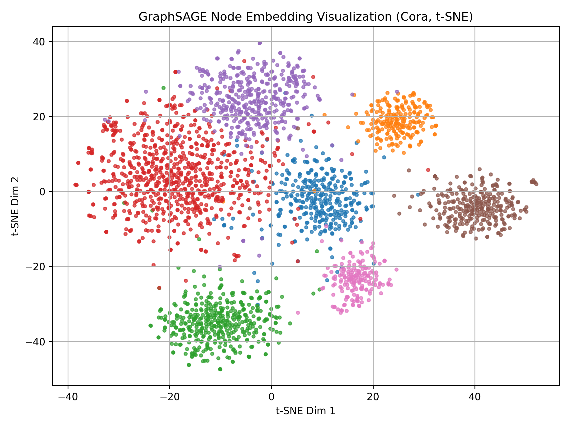}\\
		\scriptsize (d) GraphSAGE
	\end{minipage}
	
	\vspace{6pt}
	\begin{minipage}[b]{0.48\linewidth}\centering
		\panelimg{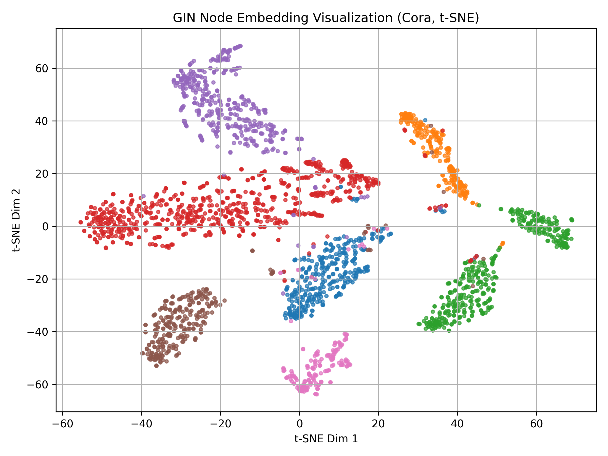}\\
		\scriptsize (e) GIN
	\end{minipage}\hfill
	\begin{minipage}[b]{0.48\linewidth}\centering
		\panelimg{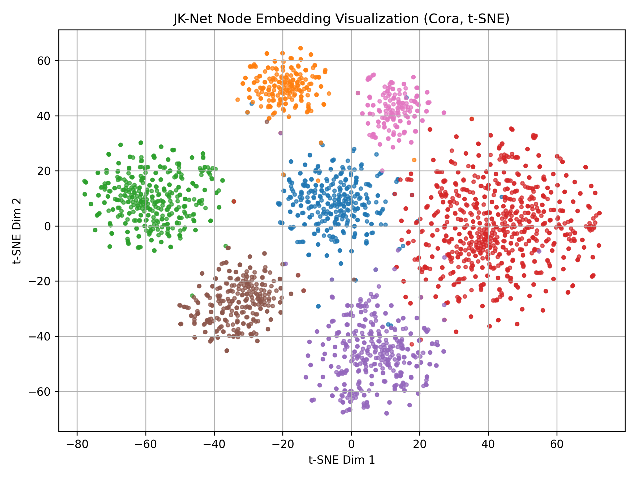}\\
		\scriptsize (f) JK-Net
	\end{minipage}
	
	\vspace{6pt}
	\begin{minipage}[b]{0.48\linewidth}\centering
		\panelimg{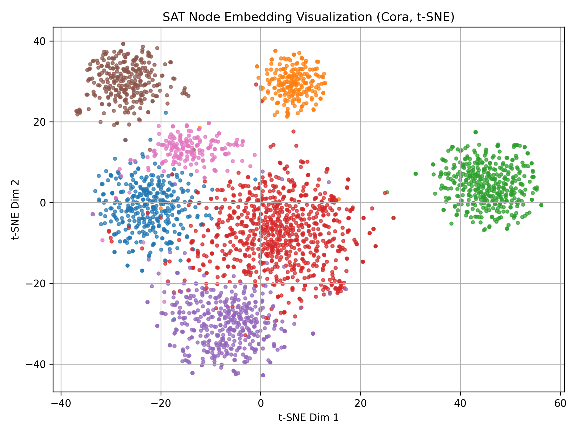}\\
		\scriptsize (g) SAT
	\end{minipage}\hfill
	\begin{minipage}[b]{0.48\linewidth}\centering
		\panelimg{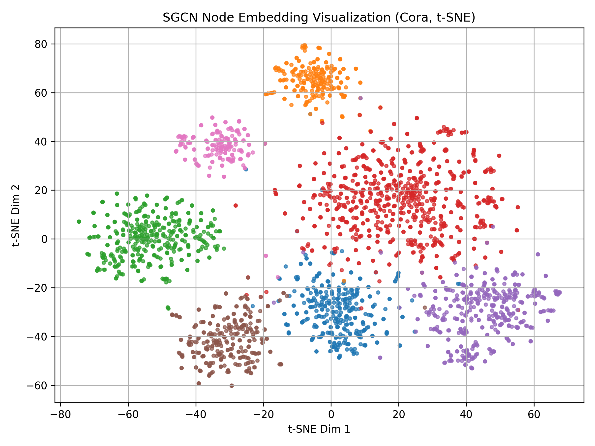}\\
		\scriptsize (h) SGCN
	\end{minipage}
	
	\caption{Node representation visualization on the Cora dataset (Part 1/2).}
	\label{fig:cora-tsne-a}
\end{figure}

\begin{figure}[p]  
	\centering
	\newcommand{\panelimgB}[1]{\includegraphics[width=\linewidth,height=0.21\textheight,keepaspectratio]{#1}\vspace{2pt}}
	
	\begin{minipage}[b]{0.48\linewidth}\centering
		\panelimgB{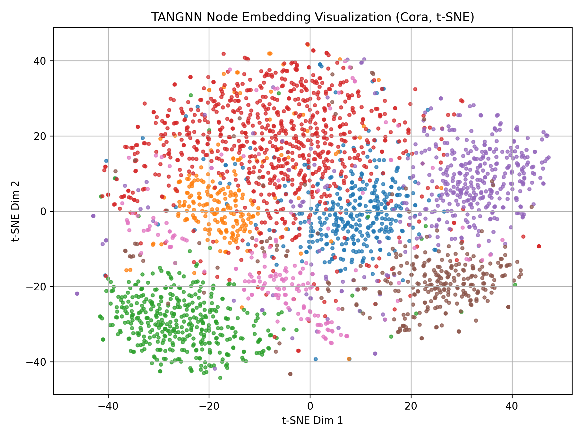}\\
		\scriptsize (i) TANGNN
	\end{minipage}\hfill
	\begin{minipage}[b]{0.48\linewidth}\centering
		\panelimgB{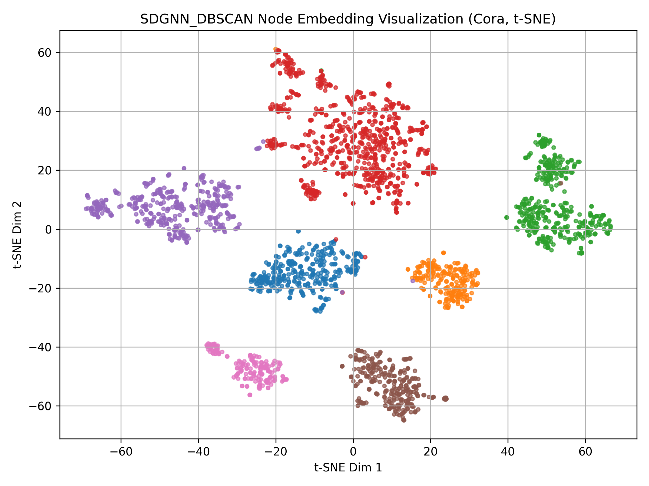}\\
		\scriptsize (j) SDGNN-DBSCAN
	\end{minipage}
	
	\vspace{6pt}
	\begin{minipage}[b]{0.48\linewidth}\centering
		\panelimgB{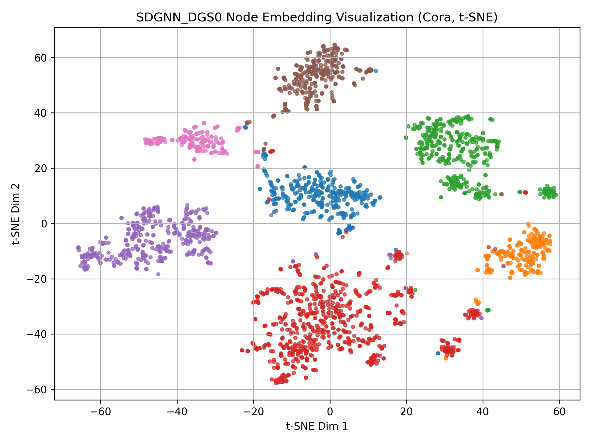}\\
		\scriptsize (k) SDGNN-DGS0
	\end{minipage}\hfill
	\begin{minipage}[b]{0.48\linewidth}\centering
		\panelimgB{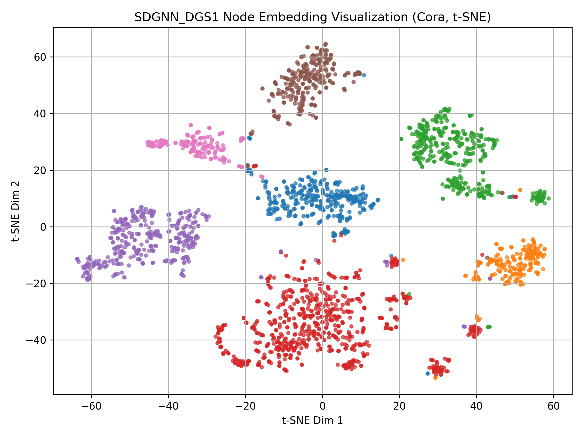}\\
		\scriptsize (l) SDGNN-DGS1
	\end{minipage}
	
	\vspace{6pt}
	\begin{minipage}[b]{0.48\linewidth}\centering
		\panelimgB{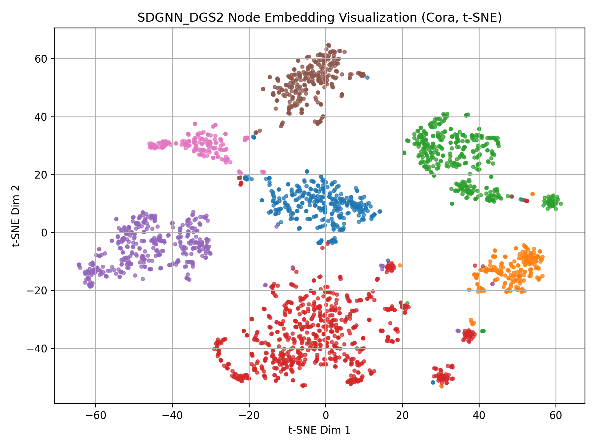}\\
		\scriptsize (m) SDGNN-DGS2
	\end{minipage}\hfill
	\begin{minipage}[b]{0.48\linewidth}\centering
		\panelimgB{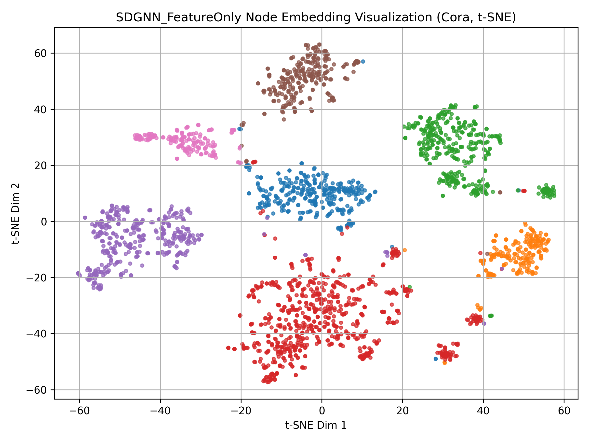}\\
		\scriptsize (n) SDGNN-FeatureOnly
	\end{minipage}
	
	\vspace{6pt}
	\begin{minipage}[b]{0.48\linewidth}\centering
		\panelimgB{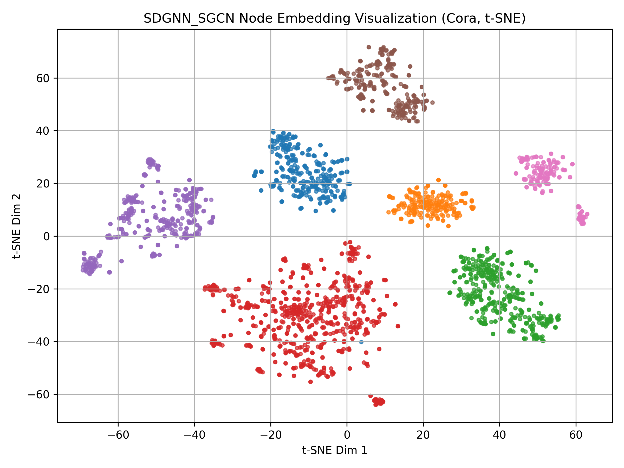}\\
		\scriptsize (o) SDGNN-SGCN
	\end{minipage}
	
	\caption{Node representation visualization on the Cora dataset (Part 2/2).}
	\label{fig:cora-tsne-b}
\end{figure}

The visualization results show that the SDGNN series outperforms traditional parameterized GNNs in terms of cluster aggregation quality and semantic separability. Among them, SDGNN-DBSCAN forms tight and uniform clusters through density-aware neighborhood partitioning, effectively distinguishing boundary nodes from bridge nodes, especially in graphs with significant differences in neighborhood density. This method is suitable for complex structures and networks with strong community heterogeneity. SDGNN-DGS0 maintains cluster compactness while showing smooth transitions at inter-class boundaries, reflecting its good adaptability to diverse structures and feature distributions. It is the most balanced solution among the four strategies. SDGNN-FeatureOnly, by completely discarding structural constraints, tends to organize nodes based on semantic similarity. Its inter-class gaps are more pronounced, but some sparse regions show isolated node distributions. In contrast, SDGNN-SGCN, relying on normalized multi-hop propagation to capture global topological relations, produces more comprehensive cluster distributions. Although the sharpness of inter-class boundaries is somewhat reduced, it performs stably in tasks where global consistency is highly required.

In comparison to baseline models, traditional parameterized GNNs (such as GCN, GAT, JK-Net) generally exhibit ``over-smoothing'' in the embedding space: nodes of the same class have ambiguous boundaries, and some categories even overlap in their distributions, especially at category boundary nodes, where discrimination ability is insufficient. Models like GraphSAGE and SAT, which introduce structure-adaptive mechanisms, alleviate this issue to some extent. However, because they rely on trainable parameters, the clusters they generate remain loose. ComBSAGE and TANGNN show improvements in alleviating over-smoothing, but their intra-class node aggregation is still inferior to the SDGNN series, and the representations of cross-community nodes tend to be more discrete.

Overall, the SDGNN series significantly outperforms comparison models in terms of intra-class compactness, inter-class separability, and visualization clarity of node representations. Among them, SDGNN-DBSCAN is better suited for handling complex networks with significant community heterogeneity; SDGNN-DGS0 achieves the best balance between efficiency and representation quality; SDGNN-FeatureOnly emphasizes semantic consistency; and SDGNN-SGCN provides global topological features with the lowest computational cost.

\subsection{Interdisciplinary Experiment}
\label{sec:interdisc}

To further evaluate the applicability and robustness of SDGNN and its variants in cross-domain, multi-modal tasks, we designed an interdisciplinary node classification experiment. Unlike the conventional classification tasks mentioned earlier, interdisciplinary tasks require models to handle scientific literature networks that have both complex citation network structures and high-dimensional semantic features, while accurately distinguishing between single-disciplinary and interdisciplinary nodes. Since the traditional PubMed dataset label system only provides a single topic category, making it unsuitable for interdisciplinary research, we performed multiple rounds of cleaning and reconstruction based on the original dataset \cite{ref35} to build a citation network suitable for this study.

In the data preprocessing stage, the original graph contained 2,846,599 nodes. To ensure the reliability and statistical sufficiency of the experiment, we selected only literature published in 2020 to construct the subgraph, with a total of 38,931 nodes. This selection avoids the impact of missing metadata in early literature and the issue of insufficient statistics due to a low number of recent publications. Then, based on the titles and abstracts of each paper, we used the SPECTER \cite{ref36} pre-trained language model to extract $768$-dimensional semantic embedding vectors as the high-dimensional text features for the nodes. For the labeling system, we used MeSH Headings \cite{ref37} to select the node topics: if a paper has only one topic, it is labeled as a single-disciplinary node; if it has more than one topic, it is labeled as an interdisciplinary node, thus capturing the cross-domain attributes of the literature. Finally, based on the citation relationships between nodes, we constructed an undirected sparse graph, resulting in a citation network that combines both complex network structure and high-dimensional semantic features.

On the preprocessed PubMed graph, we conducted a systematic evaluation of all SDGNN variants. The overall experimental process is the same as that of the node classification task described earlier. First, different group partitioning strategies are used to divide the communities or clusters; then, mean pooling is applied within each group to extract the aggregated features of the group; next, max pooling is performed across groups to retain the most significant cross-group feature signals. Finally, the multi-layer representations are fused through the Jumping Knowledge mechanism and input to a two-layer MLP classifier for training and prediction, thus maintaining efficiency while integrating local structural information, global feature associations, and multi-modal semantic modeling capabilities.

In interdisciplinary tasks, the domain attributes of journals often significantly affect the model's classification performance. Some journals are highly concentrated in specific subject areas, with citation networks and semantic features that are relatively homogeneous. In contrast, other journals are more interdisciplinary, covering multiple research directions, thus exhibiting greater heterogeneity in both structural connectivity and feature distribution. To explore the impact of this difference on model performance, we divided the data by journal categories, covering a total of 745 journals. Due to the large number of journals and significant distribution differences, we selected four representative journals for in-depth study: \textit{Aging (Albany NY)}, \textit{Ecotoxicology}, \textit{Health Qual Life Outcomes}, and \textit{Oecologia}. These journals differ in node size, network connectivity, and label range. Some of the journals have fewer edges, primarily because their citation relationships are more likely to cross into other journals rather than form dense edges within the same journal, making them structurally closer to the form of a ``cross-community sparse network''. The specific information for the four groups of journals is summarized in Table~\ref{tab:journal-stats}.

\begin{table}[ht]
	\centering
	\caption{Specific information of different journals.}
	\label{tab:journal-stats}
	\footnotesize                             
	\setlength{\tabcolsep}{4pt}               
	\begin{tabular}{p{0.30\linewidth} c c c}  
		\hline
		\textbf{Journal Name} & \textbf{Nodes} & \textbf{Edges} & \textbf{Categories} \\
		\hline
		Aging (Albany NY)         & 285 & 26  & 1--8 \\
		Ecotoxicology             & 381 & 109 & 1--6 \\
		Health Qual Life Outcomes & 376 & 10  & 1--5 \\
		Oecologia                 & 289 & 12  & 1--7 \\
		\hline
	\end{tabular}
\end{table}

To further evaluate the performance of SDGNN and its variants across different structural environments, we conducted comparison experiments with multiple mainstream graph neural network methods on four journal subgraphs, including parameterized methods (such as GAT, GCN, GIN, GraphSAGE, etc.) and various non-parameterized variants of SDGNN. The experimental process for each model is consistent with the interdisciplinary node classification task described earlier, ensuring comparability of the results. The experimental results are summarized in Tables~\ref{tab:aging}--\ref{tab:oecologia}.

\begin{table*}[ht]
	\centering
	\caption{Classification accuracy summary for different methods on the \emph{Aging (Albany NY)} journal. The best results in each column are highlighted in bold.}
	\label{tab:aging}
	\resizebox{\textwidth}{!}{%
		\begin{tabular}{lccccccccc}
			\hline
			\textbf{Method} & \textbf{5\%} & \textbf{15\%} & \textbf{25\%} & \textbf{35\%} & \textbf{45\%} & \textbf{55\%} & \textbf{65\%} & \textbf{75\%} & \textbf{85\%} \\
			\hline
			ComBSAGE & 0.5449 & 0.5749 & 0.6022 & 0.6000 & 0.6295 & 0.6178 & 0.5944 & 0.5591 & 0.6133 \\
			GAT & 0.5588 & 0.6037 & 0.5817 & 0.6063 & 0.6109 & 0.5941 & 0.6194 & 0.5636 & 0.6133 \\
			GCN & 0.5630 & 0.5786 & 0.5806 & 0.5861 & 0.5891 & 0.6099 & 0.5583 & 0.5909 & 0.4800 \\
			GIN & 0.5868 & 0.5414 & 0.5925 & 0.5949 & 0.5767 & 0.6059 & 0.6167 & 0.6136 & 0.5067 \\
			GraphSAGE & 0.5753 & 0.5842 & 0.5914 & 0.5835 & 0.6093 & 0.6040 & 0.6139 & 0.6318 & 0.6400 \\
			JK-NET & 0.5844 & 0.5833 & 0.5785 & 0.5861 & 0.6047 & 0.6198 & 0.5667 & 0.6318 & 0.6133 \\
			SAT & 0.5877 & 0.5870 & 0.5935 & 0.5595 & 0.6124 & 0.5921 & 0.5944 & 0.6545 & 0.6000 \\
			SGCN & 0.5778 & 0.6056 & 0.5624 & 0.6101 & 0.5845 & 0.5644 & 0.6056 & 0.6136 & 0.5067 \\
			TANGNN & 0.5844 & 0.6037 & 0.6161 & 0.5633 & 0.5550 & 0.6079 & 0.5889 & 0.6000 & 0.6000 \\
			SDGNN\_DBSCAN & 0.6049 & 0.6009 & \textbf{0.6280} & 0.6253 & 0.6047 & 0.6317 & 0.6139 & 0.6500 & 0.5867 \\
			SDGNN\_DGS0 & 0.5827 & 0.6112 & 0.6194 & 0.6392 & 0.6202 & \textbf{0.6337} & 0.6472 & 0.6000 & 0.6133 \\
			SDGNN\_DGS1 & 0.5671 & \textbf{0.6186} & 0.5839 & 0.6152 & \textbf{0.6372} & \textbf{0.6337} & \textbf{0.6583} & 0.5500 & 0.6133 \\
			SDGNN\_DGS2 & 0.5621 & 0.6056 & 0.6140 & 0.6139 & 0.6264 & 0.6257 & 0.5917 & 0.6318 & \textbf{0.7333} \\
			SDGNN\_FeatureOnly & 0.5663 & 0.5795 & 0.6129 & 0.5975 & 0.6124 & 0.6257 & 0.6194 & \textbf{0.6636} & 0.6267 \\
			SDGNN\_SGCN & \textbf{0.6206} & 0.5833 & 0.6247 & \textbf{0.6405} & 0.6357 & 0.6158 & 0.5806 & \textbf{0.6636} & 0.5733 \\
			\hline
	\end{tabular}}%
\end{table*}

\begin{table*}[ht]
	\centering
	\caption{Classification accuracy summary for different methods on the \emph{Ecotoxicology} journal. The best results in each column are highlighted in bold.}
	\label{tab:ecotox}
	\resizebox{\textwidth}{!}{%
		\begin{tabular}{lccccccccc}
			\hline
			\textbf{Method} & \textbf{5\%} & \textbf{15\%} & \textbf{25\%} & \textbf{35\%} & \textbf{45\%} & \textbf{55\%} & \textbf{65\%} & \textbf{75\%} & \textbf{85\%} \\
			\hline
			ComBSAGE & 0.6238 & 0.6371 & 0.6782 & 0.6590 & 0.6756 & 0.6687 & 0.6875 & 0.6793 & 0.6900 \\
			GAT & 0.6105 & 0.6000 & 0.6427 & 0.6638 & 0.6535 & 0.6896 & 0.6625 & 0.6103 & 0.6600 \\
			GCN & 0.6086 & 0.6007 & 0.6298 & 0.6457 & 0.6326 & 0.6619 & 0.6750 & 0.6897 & 0.7200 \\
			GIN & 0.6299 & {0.6685} & 0.6774 & 0.6648 & 0.6709 & 0.6896 & 0.6729 & 0.6655 & 0.6600 \\
			GraphSAGE & 0.6206 & 0.6210 & 0.6573 & 0.6124 & 0.6326 & 0.6448 & 0.6563 & 0.6345 & 0.7500 \\
			JK-NET & 0.6130 & 0.6336 & 0.6460 & 0.6562 & 0.6593 & 0.6776 & 0.6833 & 0.6897 & 0.6700 \\
			SAT & 0.6219 & 0.6427 & 0.6387 & 0.6438 & 0.6616 & 0.6881 & 0.6375 & 0.6448 & 0.6800 \\
			SGCN & 0.5883 & 0.6273 & 0.6565 & 0.6714 & 0.6512 & 0.6836 & 0.6771 & 0.6759 & 0.6900 \\
			TANGNN & 0.6074 & 0.6632 & 0.6548 & 0.6581 & 0.6674 & 0.6448 & 0.6688 & 0.7241 & 0.7000 \\
			SDGNN\_DBSCAN & 0.6074 & 0.6357 & 0.6573 & 0.6686 & 0.6849 & 0.6791 & 0.6708 & 0.7103 & 0.6800 \\
			SDGNN\_DGS0 & 0.6333 & 0.6266 & \textbf{0.6793} & 0.6581 & 0.6802 & 0.6866 & 0.6500 & 0.6586 & 0.6300 \\
			SDGNN\_DGS1 & 0.5827 & 0.6280 & 0.6734 & 0.6705 & 0.6884 & 0.6985 & 0.6708 & 0.6724 & 0.6300 \\
			SDGNN\_DGS2 & 0.6216 & 0.6287 & 0.6266 & \textbf{0.6752} & 0.6523 & 0.6761 & \textbf{0.7000} & 0.6793 & 0.7400 \\
			SDGNN\_FeatureOnly & 0.6278 & \textbf{0.6687} & 0.6274 & 0.6686 & \textbf{0.6965} & \textbf{0.7015} & 0.6875 & \textbf{0.7414} & \textbf{0.7800} \\
			SDGNN\_SGCN & \textbf{0.6420} & 0.6594 & 0.6621 & 0.6667 & 0.6930 & 0.6881 & 0.6792 & 0.7000 & 0.7100 \\
			\hline
	\end{tabular}}%
\end{table*}

\begin{table*}[ht]
	\centering
	\caption{Classification accuracy summary for different methods on the \emph{Health Qual Life Outcomes} journal. The best results in each column are highlighted in bold.}
	\label{tab:hqlo}
	\resizebox{\textwidth}{!}{%
		\begin{tabular}{lccccccccc}
			\hline
			\textbf{Method} & \textbf{5\%} & \textbf{15\%} & \textbf{25\%} & \textbf{35\%} & \textbf{45\%} & \textbf{55\%} & \textbf{65\%} & \textbf{75\%} & \textbf{85\%} \\
			\hline
			ComBSAGE & 0.5614 & 0.5456 & 0.6073 & 0.5894 & 0.5824 & 0.6241 & 0.5811 & 0.5614 & 0.5700 \\
			GAT & 0.5607 & 0.5498 & 0.6049 & 0.6048 & 0.6200 & 0.6105 & 0.6147 & 0.6246 & 0.5400 \\
			GCN & 0.5595 & 0.5534 & 0.5992 & 0.6010 & 0.5859 & 0.6150 & 0.6211 & 0.6246 & 0.6100 \\
			GIN & 0.4822 & 0.5717 & 0.5845 & 0.5529 & 0.5800 & 0.5729 & 0.5726 & 0.6456 & 0.5900 \\
			GraphSAGE & 0.5788 & 0.5788 & 0.5616 & 0.5817 & 0.5765 & 0.6195 & 0.5453 & 0.6632 & 0.6200 \\
			JK-NET & 0.5240 & 0.5541 & 0.5739 & 0.5846 & 0.6176 & 0.6291 & 0.6253 & 0.6386 & 0.6000 \\
			SAT & 0.5302 & 0.5696 & 0.5747 & 0.5750 & 0.5600 & 0.6120 & 0.6147 & 0.6351 & 0.6500 \\
			SGCN & 0.5664 & 0.5385 & 0.6016 & 0.5750 & 0.5824 & 0.6150 & 0.6105 & 0.6561 & 0.6000 \\
			TANGNN & 0.5065 & 0.5039 & 0.5788 & 0.5721 & 0.5694 & 0.5669 & 0.5705 & 0.5544 & 0.5900 \\
			SDGNN\_DBSCAN & 0.5632 & 0.5633 & 0.5951 & 0.6144 & 0.6129 & 0.6195 & 0.5726 & 0.6000 & 0.6700 \\
			SDGNN\_DGS0 & \textbf{0.5913} & \textbf{0.6071} & \textbf{0.6135} & 0.5981 & 0.5859 & \textbf{0.6361} & 0.6484 & 0.6386 & 0.5200 \\
			SDGNN\_DGS1 & 0.5526 & 0.5774 & 0.5959 & 0.6058 & \textbf{0.6341} & 0.5925 & 0.5832 & \textbf{0.6772} & 0.6600 \\
			SDGNN\_DGS2 & 0.5383 & 0.5880 & 0.5910 & 0.5567 & 0.5976 & 0.6256 & 0.6168 & 0.5684 & 0.5800 \\
			SDGNN\_FeatureOnly & 0.5514 & 0.5901 & 0.5608 & \textbf{0.6337} & 0.5882 & 0.5669 & 0.5432 & 0.5895 & \textbf{0.6900} \\
			SDGNN\_SGCN & 0.5396 & 0.5951 & 0.5796 & {0.6173} & 0.6235 & 0.5774 & \textbf{0.6379} & 0.6246 & {0.6500} \\
			\hline
	\end{tabular}}%
\end{table*}

\begin{table*}[ht]
	\centering
	\caption{Classification accuracy summary for different methods on the \emph{Oecologia} journal. The best results in each column are highlighted in bold.}
	\label{tab:oecologia}
	\resizebox{\textwidth}{!}{%
		\begin{tabular}{lccccccccc}
			\hline
			\textbf{Method} & \textbf{5\%} & \textbf{15\%} & \textbf{25\%} & \textbf{35\%} & \textbf{45\%} & \textbf{55\%} & \textbf{65\%} & \textbf{75\%} & \textbf{85\%} \\
			\hline
			ComBSAGE & 0.8073 & 0.8380 & 0.8328 & 0.8363 & 0.8366 & 0.8214 & 0.8432 & 0.8000 & 0.8250 \\
			GAT & 0.8178 & 0.8046 & 0.8106 & 0.8100 & 0.8183 & 0.8311 & 0.8378 & 0.8711 & 0.8500 \\
			GCN & 0.7968 & 0.7642 & 0.8138 & 0.8100 & 0.8214 & 0.8233 & 0.8243 & 0.8356 & 0.8375 \\
			GIN & 0.8200 & 0.8303 & 0.8339 & 0.8338 & 0.8229 & 0.8291 & 0.8027 & 0.8622 & 0.8125 \\
			GraphSAGE & 0.7579 & 0.7927 & 0.8222 & 0.8163 & 0.8198 & 0.8194 & 0.8027 & 0.8444 & 0.8375 \\
			JK-NET & 0.7814 & 0.8257 & 0.8328 & 0.8113 & 0.8275 & \textbf{0.8505} & 0.8189 & 0.7956 & 0.8125 \\
			SAT & \textbf{0.8243} & 0.8284 & 0.8127 & 0.8375 & 0.8229 & 0.8272 & 0.8189 & 0.8533 & 0.8500 \\
			SGCN & 0.8000 & 0.8229 & 0.8307 & 0.8418 & 0.8305 & 0.8311 & 0.8243 & 0.8000 & 0.8375 \\
			TANGNN & 0.8154 & 0.8294 & 0.8434 & 0.8363 & 0.8504 & 0.8194 & 0.8351 & 0.8444 & 0.9125 \\
			SDGNN\_DBSCAN & 0.7587 & 0.8211 & 0.8106 & 0.8213 & 0.8321 & 0.8369 & 0.8351 & 0.8178 & 0.8375 \\
			SDGNN\_DGS0 & 0.7919 & 0.8183 & 0.8127 & 0.8138 & 0.8321 & 0.8608 & 0.8081 & 0.8222 & \textbf{0.9375} \\
			SDGNN\_DGS1 & 0.7611 & 0.7945 & 0.8190 & 0.8313 & \textbf{0.8527} & 0.8291 & 0.8432 & 0.8178 & 0.8125 \\
			SDGNN\_DGS2 & 0.8243 & 0.8101 & \textbf{0.8457} & 0.8150 & 0.8260 & 0.8408 & 0.8432 & 0.8311 & 0.9000 \\
			SDGNN\_FeatureOnly & 0.7854 & \textbf{0.8381} & 0.8138 & \textbf{0.8425} & 0.8382 & 0.8369 & \textbf{0.8703} & \textbf{0.8711} & 0.8625 \\
			SDGNN\_SGCN & 0.8154 & 0.8193 & 0.7894 & 0.8138 & 0.8382 & 0.8311 & 0.8270 & 0.8533 & 0.7625 \\
			\hline
	\end{tabular}}%
\end{table*}

From Tables~\ref{tab:aging}--\ref{tab:oecologia}, it can be seen that despite the significant differences in node scale, citation density, and label distribution across the four journals, SDGNN and its various aggregation variants consistently exhibit high stability and adaptability under different levels of supervision. For subgraphs with sparse edges like \emph{Aging (Albany NY)} and \emph{Health Qual Life Outcomes}, traditional parameterized models are generally limited under low supervision conditions. However, SDGNN-SGCN, SDGNN-DBSCAN, and some DGS-based variants remain competitive, indicating their stronger structural compensation and cross-node information integration capabilities when information is scarce. In the more densely connected \emph{Ecotoxicology} dataset, the overall accuracy level of the models improves. The feature-driven SDGNN-FeatureOnly performs particularly well in medium to high supervision conditions, showing that weakening structural constraints can help improve semantic capture efficiency when high-dimensional semantic features are abundant and structural connections are rich. In subgraphs with a wider label range, such as \emph{Oecologia}, the SDGNN-DGS2 based on feature similarity clustering and the purely feature-driven FeatureOnly both achieve the highest accuracy multiple times, reflecting the advantages of these strategies in scenarios with multi-label differentiation.

In the cross-group aggregation process, we observed significant differences between interdisciplinary tasks and conventional node classification tasks. In conventional tasks, max pooling is effective in quickly highlighting dominant structural or semantic signals by retaining the strongest feature responses within each group, which is particularly suitable for single-domain networks where decisions are supported by a single dominant group. However, in interdisciplinary tasks, the cross-domain attributes of nodes often result from the combined contributions of multiple academic groups. If only the strongest signals are taken, other important but weaker information may be overlooked, affecting the accurate representation of cross-domain features.

Based on this observation, we introduced mean pooling as a supplementary solution in the cross-disciplinary experiments. In the cross-group fusion stage, mean pooling performs equal-weight averaging of features from different groups, thus preserving the overall contribution of each group. This design is highly consistent with structural diversity theory, which emphasizes the comprehensive coverage of all independent structural branches in a neighborhood, rather than focusing solely on the strongest single signal. In practice, we first perform community or cluster partitioning based on different group partitioning strategies, then apply mean pooling within groups, and perform mean pooling again between groups to fully integrate multi-source information. This method captures the complementary features of cross-domain nodes more comprehensively, particularly when multiple sources of information define the node's semantics.

To fairly compare the adaptability of the two strategies, we evaluated two cross-group fusion schemes under the same group partitioning settings: \emph{mean+max} (mean pooling within groups + max pooling between groups) and \emph{mean+mean} (mean pooling within groups + mean pooling between groups). The systematic evaluation was conducted on the four journal subgraphs, reported in Tables~\ref{tab:aging-fusion}--\ref{tab:oecologia-fusion}.

\begin{table*}[ht]
	\centering
	\caption{Classification accuracy summary for different strategies on the \emph{Aging (Albany NY)} journal. The best results in each column are highlighted in bold.}
	\label{tab:aging-fusion}
	\resizebox{\textwidth}{!}{%
		\begin{tabular}{lccccccccc}
			\hline
			& \textbf{5\%} & \textbf{15\%} & \textbf{25\%} & \textbf{35\%} & \textbf{45\%} & \textbf{55\%} & \textbf{65\%} & \textbf{75\%} & \textbf{85\%} \\
			\hline
			\multicolumn{10}{l}{\textit{mean+mean (within-group mean, cross-group mean)}} \\
			SDGNN\_DBSCAN      & 0.5877 & \textbf{0.6515} & 0.6280 & \textbf{0.6917} & 0.6326 & 0.6733 & 0.6325 & 0.6250 & 0.5867 \\
			SDGNN\_DGS0        & 0.6074 & 0.6260 & 0.6194 & 0.6228 & 0.6543 & \textbf{0.6871} & 0.6361 & 0.6182 & 0.6400 \\
			SDGNN\_DGS1        & 0.6074 & \textbf{0.6515} & 0.6280 & \textbf{0.6917} & \textbf{0.6543} & \textbf{0.6871} & 0.6361 & 0.6250 & 0.6400 \\
			SDGNN\_DGS2        & 0.5473 & 0.6009 & \textbf{0.6323} & 0.6456 & 0.6419 & 0.6653 & 0.6472 & 0.6000 & 0.5867 \\
			SDGNN\_FeatureOnly & 0.6058 & 0.6279 & 0.6011 & 0.6266 & 0.6155 & 0.6416 & {0.6500} & 0.6136 & {0.6667} \\
			SDGNN\_SGCN        & \textbf{0.6206} & 0.5647 & 0.6161 & 0.5987 & 0.6372 & 0.6376 & 0.6306 & \textbf{0.6682} & 0.6133 \\
			\hline
			\multicolumn{10}{l}{\textit{mean+max (within-group mean, cross-group max)}} \\
			SDGNN\_DBSCAN      & 0.6049 & 0.6009 & 0.6280 & 0.6253 & 0.6047 & 0.6317 & 0.6139 & 0.6500 & 0.5867 \\
			SDGNN\_DGS0        & 0.5827 & 0.6112 & 0.6194 & 0.6392 & 0.6202 & 0.6337 & 0.6472 & 0.6000 & 0.6133 \\
			SDGNN\_DGS1        & 0.5671 & 0.6186 & 0.5839 & 0.6152 & 0.6372 & 0.6337 & \textbf{0.6583} & 0.5500 & 0.6133 \\
			SDGNN\_DGS2        & 0.5621 & 0.6056 & 0.6140 & 0.6139 & 0.6264 & 0.6257 & 0.5917 & 0.6318 & \textbf{0.7333} \\
			SDGNN\_FeatureOnly & 0.5663 & 0.5795 & 0.6129 & 0.5975 & 0.6124 & 0.6257 & 0.6194 & {0.6636} & 0.6267 \\
			SDGNN\_SGCN        & \textbf{0.6206} & 0.5833 & 0.6247 & 0.6405 & 0.6357 & 0.6158 & 0.5806 & {0.6636} & 0.5733 \\
			\hline
	\end{tabular}}%
\end{table*}

\begin{table*}[ht]
	\centering
	\caption{Classification accuracy summary for different strategies on the \emph{Ecotoxicology} journal. The best results in each column are highlighted in bold.}
	\label{tab:ecotox-fusion}
	\resizebox{\textwidth}{!}{%
		\begin{tabular}{lccccccccc}
			\hline
			& \textbf{5\%} & \textbf{15\%} & \textbf{25\%} & \textbf{35\%} & \textbf{45\%} & \textbf{55\%} & \textbf{65\%} & \textbf{75\%} & \textbf{85\%} \\
			\hline
			\multicolumn{10}{l}{\textit{mean+mean (within-group mean, cross-group mean)}} \\
			SDGNN\_DBSCAN      & 0.6130 & 0.6406 & 0.6637 & 0.6667 & 0.6756 & 0.6955 & 0.6813 & 0.6966 & 0.6600 \\
			SDGNN\_DGS0        & {0.6519} & {0.6629} & 0.6661 & 0.6800 & 0.6756 & 0.6806 & \textbf{0.6979} & 0.6931 & 0.7400 \\
			SDGNN\_DGS1        & {0.6519} & {0.6629} & 0.6661 & 0.6800 & 0.6756 & 0.6955 & \textbf{0.6979} & 0.6966 & 0.7400 \\
			SDGNN\_DGS2        & 0.6488 & 0.6587 & 0.6637 & \textbf{0.6914} & 0.6756 & 0.6731 & 0.6917 & 0.6966 & 0.6800 \\
			SDGNN\_FeatureOnly & \textbf{0.6574} & 0.6559 & \textbf{0.6710} & 0.6867 & 0.6721 & 0.7000 & 0.6813 & 0.6655 & 0.6700 \\
			SDGNN\_SGCN        & 0.5747 & 0.6420 & 0.6677 & 0.6848 & 0.6802 & 0.6642 & 0.6896 & 0.7138 & 0.7000 \\
			\hline
			\multicolumn{10}{l}{\textit{mean+max (within-group mean, cross-group max)}} \\
			SDGNN\_DBSCAN      & 0.6074 & 0.6357 & 0.6573 & 0.6686 & 0.6849 & 0.6791 & 0.6708 & 0.7103 & 0.6800 \\
			SDGNN\_DGS0        & 0.6333 & 0.6266 & 0.6793 & 0.6581 & 0.6802 & 0.6866 & 0.6500 & 0.6586 & 0.6300 \\
			SDGNN\_DGS1        & 0.5827 & 0.6280 & 0.6734 & 0.6705 & 0.6884 & 0.6985 & 0.6708 & 0.6724 & 0.6300 \\
			SDGNN\_DGS2        & 0.6216 & 0.6287 & 0.6266 & 0.6752 & 0.6523 & 0.6761 & 0.7000 & 0.6793 & 0.7400 \\
			SDGNN\_FeatureOnly & 0.6278 & \textbf{0.6687} & 0.6274 & 0.6686 & \textbf{0.6965} & \textbf{0.7015} & 0.6875 & \textbf{0.7414} & \textbf{0.7800} \\
			SDGNN\_SGCN        & 0.6420 & 0.6594 & 0.6621 & 0.6667 & 0.6930 & 0.6881 & 0.6792 & 0.7000 & 0.7100 \\
			\hline
	\end{tabular}}%
\end{table*}

\begin{table*}[ht]
	\centering
	\caption{Classification accuracy summary for different strategies on the \emph{Ecotoxicology} journal. The best results in each column are highlighted in bold.}
	\label{tab:ecotox-fusion}
	\resizebox{\textwidth}{!}{%
		\begin{tabular}{lccccccccc}
			\hline
			& \textbf{5\%} & \textbf{15\%} & \textbf{25\%} & \textbf{35\%} & \textbf{45\%} & \textbf{55\%} & \textbf{65\%} & \textbf{75\%} & \textbf{85\%} \\
			\hline
			\multicolumn{10}{l}{\textit{mean+mean (within-group mean, cross-group mean)}} \\
			SDGNN\_DBSCAN      & 0.6130 & 0.6406 & 0.6637 & 0.6667 & 0.6756 & 0.6955 & 0.6813 & 0.6966 & 0.6600 \\
			SDGNN\_DGS0        & \textbf{0.6519} & 0.6629 & 0.6661 & 0.6800 & 0.6756 & 0.6806 & \textbf{0.6979} & 0.6931 & 0.7400 \\
			SDGNN\_DGS1        & \textbf{0.6519} & 0.6629 & 0.6661 & 0.6800 & 0.6756 & 0.6955 & \textbf{0.6979} & 0.6966 & 0.7400 \\
			SDGNN\_DGS2        & 0.6488 & 0.6587 & 0.6637 & \textbf{0.6914} & 0.6756 & 0.6731 & 0.6917 & 0.6966 & 0.6800 \\
			SDGNN\_FeatureOnly & 0.6574 & 0.6559 & 0.6710 & 0.6867 & 0.6721 & 0.7000 & 0.6813 & 0.6655 & 0.6700 \\
			SDGNN\_SGCN        & 0.5747 & 0.6420 & 0.6677 & 0.6848 & 0.6802 & 0.6642 & 0.6896 & 0.7138 & 0.7000 \\
			\hline
			\multicolumn{10}{l}{\textit{mean+max (within-group mean, cross-group max)}} \\
			SDGNN\_DBSCAN      & 0.6074 & 0.6357 & 0.6573 & 0.6686 & 0.6849 & 0.6791 & 0.6708 & 0.7103 & 0.6800 \\
			SDGNN\_DGS0        & 0.6333 & 0.6266 & 0.6793 & 0.6581 & 0.6802 & 0.6866 & 0.6500 & 0.6586 & 0.6300 \\
			SDGNN\_DGS1        & 0.5827 & 0.6280 & 0.6734 & 0.6705 & 0.6884 & 0.6985 & 0.6708 & 0.6724 & 0.6300 \\
			SDGNN\_DGS2        & 0.6216 & 0.6287 & 0.6266 & 0.6752 & 0.6523 & 0.6761 & 0.7000 & 0.6793 & 0.7400 \\
			SDGNN\_FeatureOnly & 0.6278 & \textbf{0.6687} & 0.6274 & 0.6686 & \textbf{0.6965} & \textbf{0.7015} & 0.6875 & \textbf{0.7414} & \textbf{0.7800} \\
			SDGNN\_SGCN        & 0.6420 & 0.6594 & \textbf{0.6793} & 0.6667 & 0.6930 & 0.6881 & 0.6792 & 0.7000 & 0.7100 \\
			\hline
	\end{tabular}}%
\end{table*}

\begin{table*}[ht]
	\centering
	\caption{Classification accuracy summary for different strategies on the \emph{Oecologia} journal. The best results in each column are highlighted in bold.}
	\label{tab:oecologia-fusion}
	\resizebox{\textwidth}{!}{%
		\begin{tabular}{lccccccccc}
			\hline
			& \textbf{5\%} & \textbf{15\%} & \textbf{25\%} & \textbf{35\%} & \textbf{45\%} & \textbf{55\%} & \textbf{65\%} & \textbf{75\%} & \textbf{85\%} \\
			\hline
			\multicolumn{10}{l}{\textit{mean+mean (within-group mean, cross-group mean)}} \\
			SDGNN\_DBSCAN      & \textbf{0.8251} & 0.8138 & 0.8138 & {0.8363} & {0.8427} & 0.8175 & 0.8568 & 0.8311 & 0.8500 \\
			SDGNN\_DGS0        & 0.8162 & 0.8018 & 0.8296 & 0.8250 & 0.8366 & \textbf{0.8641} & 0.8514 & 0.8578 & 0.8375 \\
			SDGNN\_DGS1        & \textbf{0.8251} & 0.8138 & 0.8296 & {0.8363} & {0.8427} & \textbf{0.8641} & 0.8568 & 0.8578 & 0.8500 \\
			SDGNN\_DGS2        & 0.8211 & 0.8083 & 0.8201 & 0.8313 & 0.8412 & 0.8311 & 0.8324 & 0.8311 & 0.8375 \\
			SDGNN\_FeatureOnly & 0.8162 & 0.7945 & 0.8222 & 0.8250 & 0.8198 & 0.8311 & 0.8432 & 0.8489 & 0.8375 \\
			SDGNN\_SGCN        & 0.7781 & 0.7569 & 0.8265 & 0.8088 & 0.8305 & 0.8117 & 0.8324 & 0.7956 & 0.8750 \\
			\hline
			\multicolumn{10}{l}{\textit{mean+max (within-group mean, cross-group max)}} \\
			SDGNN\_DBSCAN      & 0.7587 & 0.8211 & 0.8106 & 0.8213 & 0.8321 & 0.8369 & 0.8351 & 0.8178 & 0.8375 \\
			SDGNN\_DGS0        & 0.7919 & 0.8183 & 0.8127 & 0.8138 & 0.8321 & 0.8608 & 0.8081 & 0.8222 & \textbf{0.9375} \\
			SDGNN\_DGS1        & 0.7611 & 0.7945 & 0.8190 & 0.8313 & \textbf{0.8527} & 0.8291 & 0.8432 & 0.8178 & 0.8125 \\
			SDGNN\_DGS2        & 0.8243 & 0.8101 & \textbf{0.8457} & 0.8150 & 0.8260 & 0.8408 & 0.8432 & 0.8311 & 0.9000 \\
			SDGNN\_FeatureOnly & 0.7854 & \textbf{0.8381} & 0.8138 & \textbf{0.8425} & 0.8382 & 0.8369 & \textbf{0.8703} & \textbf{0.8711} & 0.8625 \\
			SDGNN\_SGCN        & 0.8154 & 0.8193 & 0.7894 & 0.8138 & 0.8382 & 0.8311 & 0.8270 & 0.8533 & 0.7625 \\
			\hline
	\end{tabular}}%
\end{table*}

From the results in Tables~\ref{tab:aging-fusion}--\ref{tab:oecologia-fusion}, the two cross-group fusion strategies show complementary characteristics across different journal subgraphs. For networks with sparse connections and relatively low cross-community citation ratios (such as \emph{Aging (Albany NY)} and \emph{Health Qual Life Outcomes}), node features are often more influenced by a few strongly associated groups. In these structures, \emph{mean+max} performs better at highlighting dominant feature signals, providing stable performance under low supervision ratios. On the other hand, \emph{mean+mean}, by equally preserving the information from all groups, shows greater robustness in medium to high supervision conditions, preventing the model from over-relying on individual signals.

In contrast, for networks with strong cross-community links and more diverse semantic distributions (such as \emph{Ecotoxicology} and \emph{Oecologia}), the advantage of \emph{mean+mean} becomes more pronounced. This strategy can integrate multi-source information from multiple groups, enhancing the ability to capture cross-domain features, and outperforms \emph{mean+max} under multiple supervision ratios. Overall, \emph{mean+mean} is more suitable for scenarios with multiple coexisting group signals and significant structural diversity, while \emph{mean+max} still holds advantages in networks where information is concentrated and dominated by a few groups. This difference highlights the complementary role of different aggregation methods in adapting to structural features and semantic distributions in interdisciplinary tasks.

\section{Ablation Experiment}
\label{sec:ablation}

\subsection{Impact of Layer Depth on Performance}
\label{subsec:depth}

\begin{figure}[htbp]
	\centering
	
	\begin{minipage}[b]{0.48\linewidth}\centering
		\includegraphics[width=\linewidth]{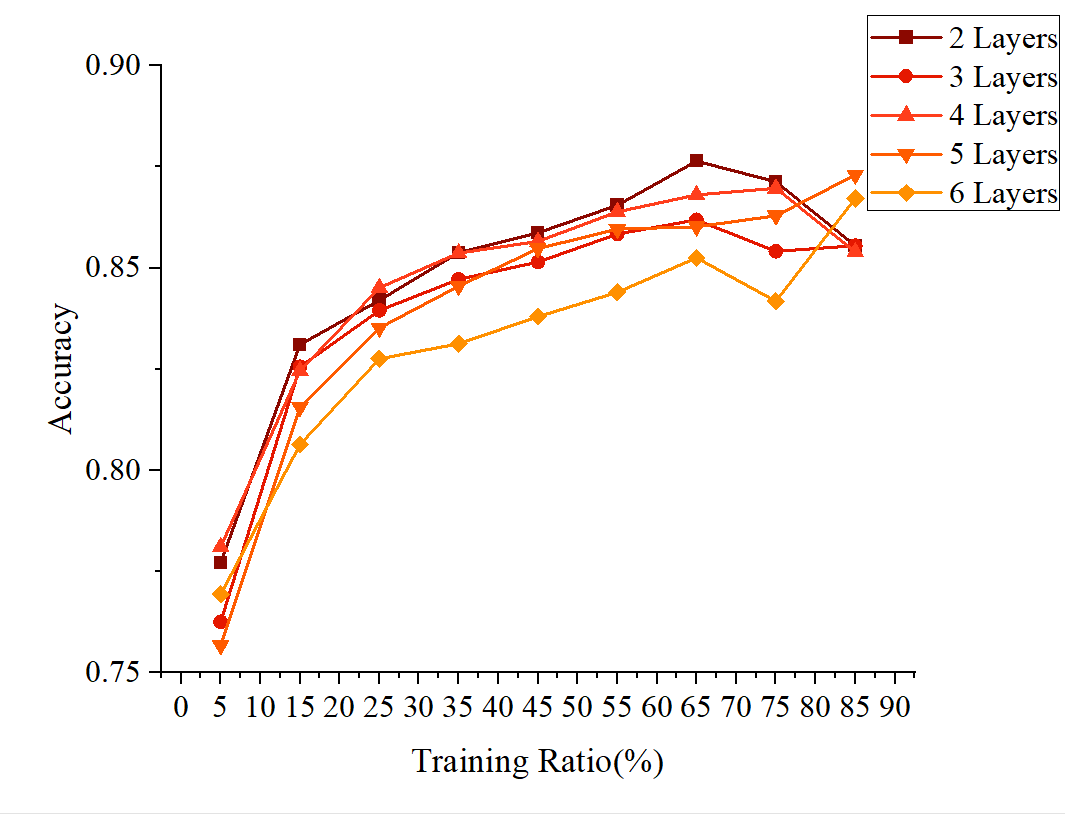}\vspace{2pt}\\
		\scriptsize (a) SDGNN-DBSCAN
	\end{minipage}\hfill
	\begin{minipage}[b]{0.48\linewidth}\centering
		\includegraphics[width=\linewidth]{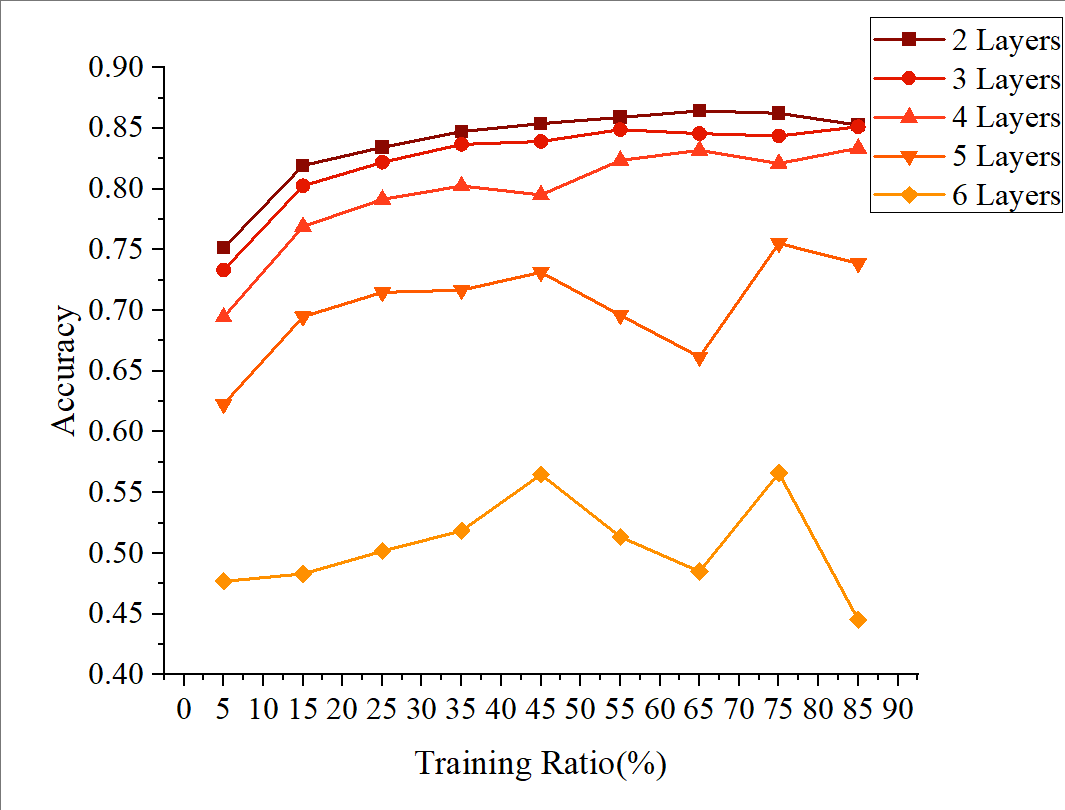}\vspace{2pt}\\
		\scriptsize (b) SDGNN-DGS0
	\end{minipage}
	
	\vspace{6pt}
	\begin{minipage}[b]{0.48\linewidth}\centering
		\includegraphics[width=\linewidth]{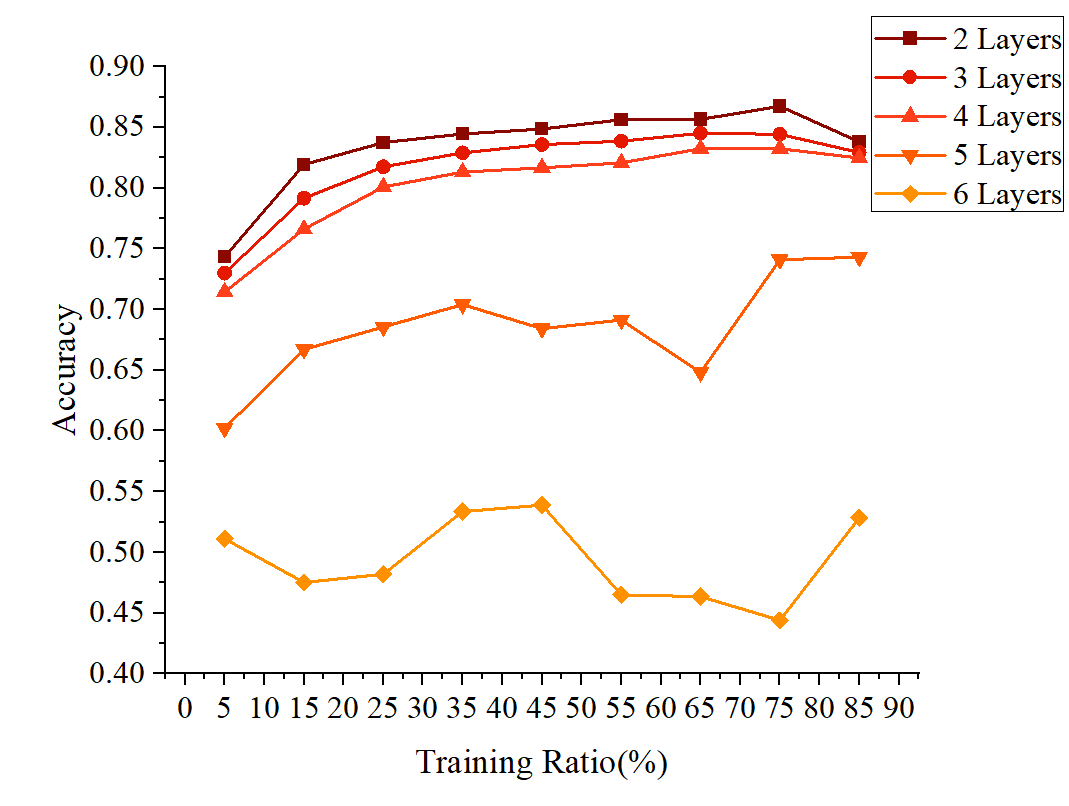}\vspace{2pt}\\
		\scriptsize (c) SDGNN-DGS1
	\end{minipage}\hfill
	\begin{minipage}[b]{0.48\linewidth}\centering
		\includegraphics[width=\linewidth]{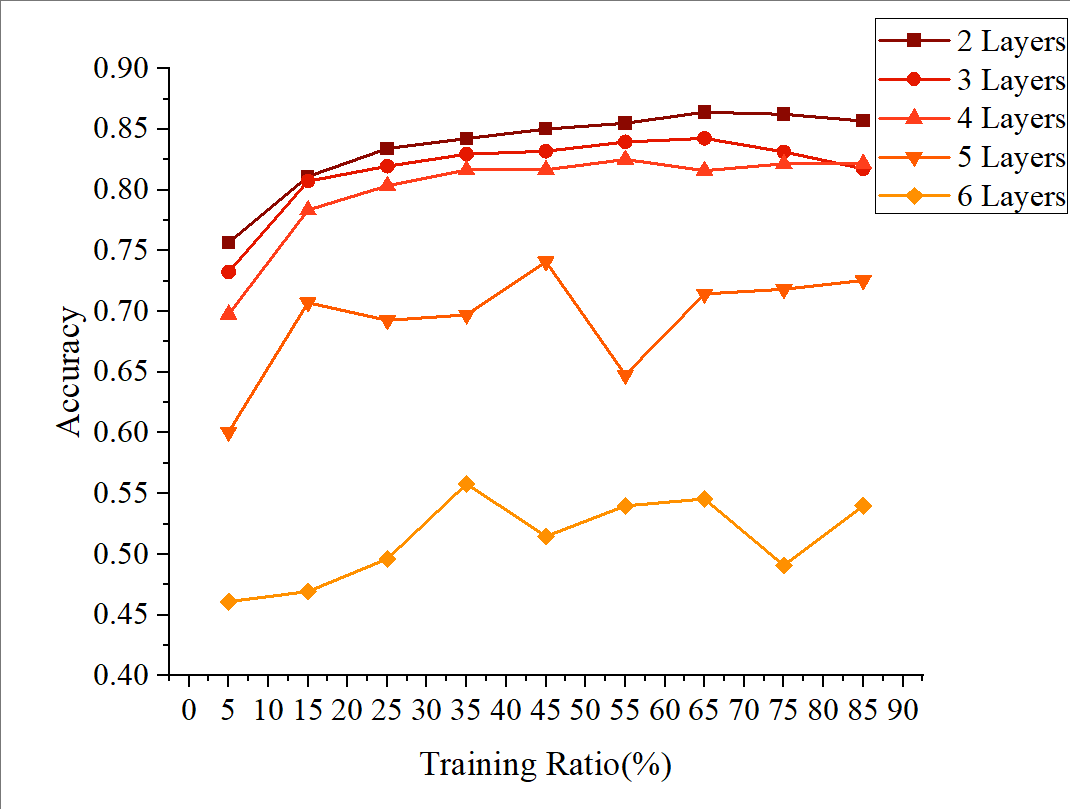}\vspace{2pt}\\
		\scriptsize (d) SDGNN-DGS2
	\end{minipage}
	
	\vspace{6pt}
	\begin{minipage}[b]{0.48\linewidth}\centering
		\includegraphics[width=\linewidth]{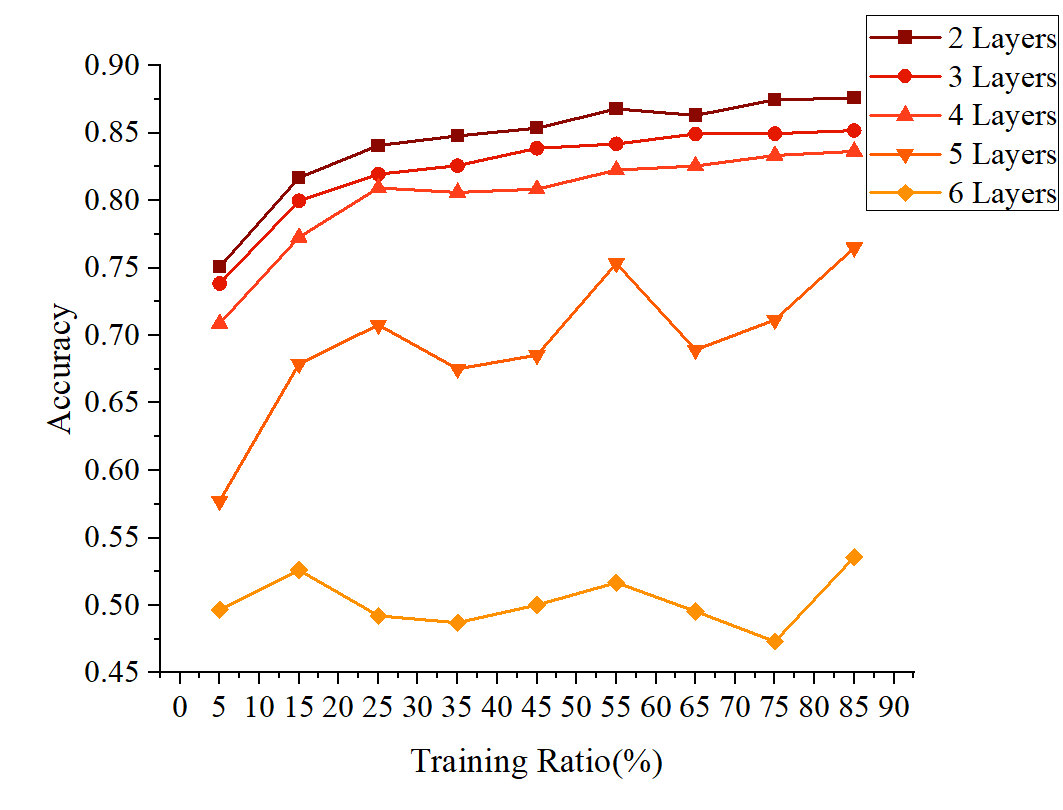}\vspace{2pt}\\
		\scriptsize (e) SDGNN-FeatureOnly
	\end{minipage}\hfill
	\begin{minipage}[b]{0.48\linewidth}\centering
		\includegraphics[width=\linewidth]{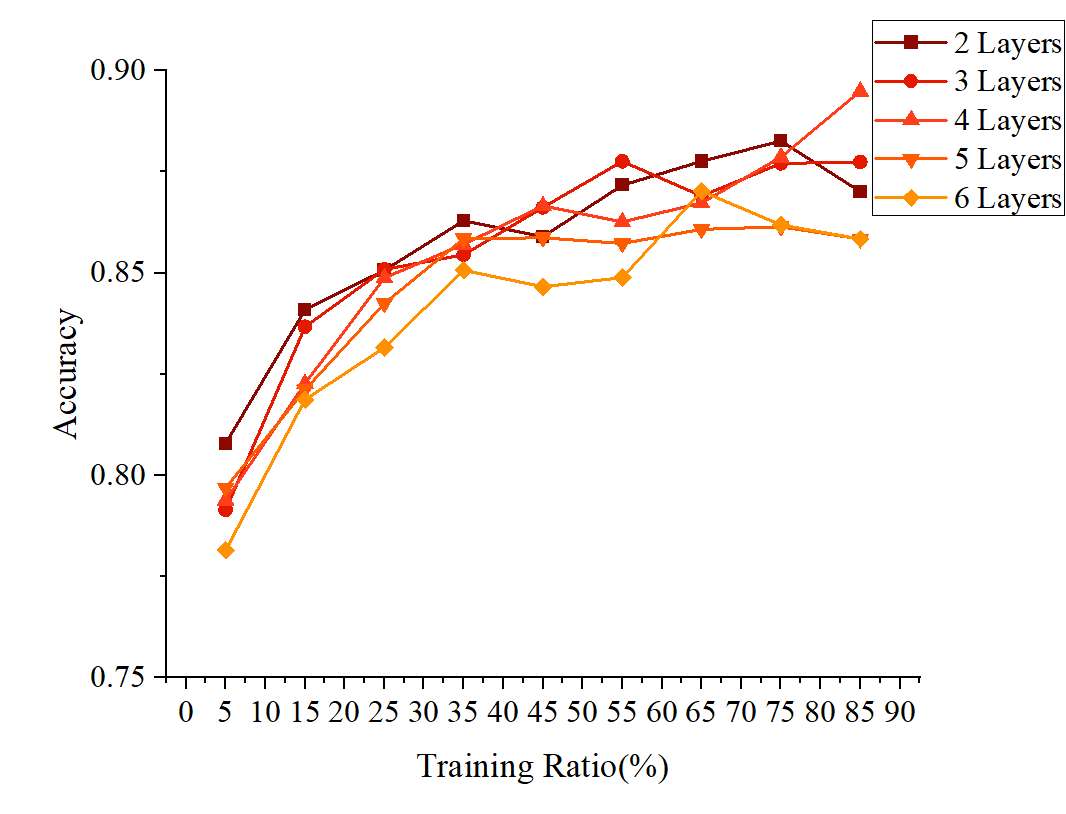}\vspace{2pt}\\
		\scriptsize (f) SDGNN-SGCN
	\end{minipage}
	
	\caption{Experiments on different aggregation strategies across depths (2--6 layers) on the Cora dataset.}
	\label{fig:cora-depth}
\end{figure}

To investigate the impact of network depth on the performance of different SDGNN strategies, we evaluated the node classification accuracy of six variants (SDGNN-DBSCAN, SDGNN-DGS0, SDGNN-DGS1, SDGNN-DGS2, SDGNN-FeatureOnly, SDGNN-SGCN) at different depths ranging from 2 to 6 layers on the Cora dataset. The results are shown in Figure~\ref{fig:cora-depth}.

Overall, all strategies achieved relatively stable performance at shallow depths (2--3 layers). However, as the depth increased, the accuracy generally decreased, with a noticeable drop in performance, especially at depths of 5--6 layers.

Specifically, SDGNN-SGCN is the most stable across all depths, reaching its peak performance at 4 layers (with the highest accuracy of 0.8949 at an 85\% training ratio), showing that global structure propagation somewhat alleviates the over-smoothing problem typically seen in deeper models. SDGNN-DBSCAN and SDGNN-DGS0 perform well in the 2--4 layer range, with slightly higher classification accuracy at 4 layers, but their performance rapidly degrades at 5 layers or more, especially under low training ratios, where the difference is more pronounced. In contrast, DGS1 and DGS2, due to the introduction of multiple iterations for updating cluster centers, maintain reasonable performance at medium depths (3--4 layers), but their accuracy significantly drops at 5--6 layers, showing the combined effects of accumulated computational errors and over-smoothing.

Overall analysis reveals that a two-layer structure strikes the best balance between performance and efficiency: on the one hand, shallow models avoid the over-smoothing problem caused by high-order neighborhood information; on the other hand, a two-layer design is sufficient to capture both first- and second-order neighborhood structure and semantic information. Therefore, in this paper, all experiments use the two-layer configuration as the standard model for SDGNN.

\subsection{Ablation Analysis}
\label{subsec:ablation-analysis}

To further evaluate the contribution of SDGNN's core components to overall performance, we conducted three sets of ablation experiments on the Cora dataset, using SDGNN-DGS0 as the representative model. The specific setups include: {Only First Layer (only1)}: Only the first layer’s neighborhood community aggregation is performed, and the dynamic partitioning aggregation of the second layer is discarded. This experiment assesses the independent contribution of the first layer’s local structure modeling. {Only Second Layer (only2)}: The first layer's community aggregation is skipped, and only the second layer's feature-driven clustering and fusion are retained. This experiment evaluates the isolated effect of higher-order semantic information. {No Jumping Knowledge (NoJK)}: Both aggregation layers are preserved, but multi-layer representations are not concatenated. Instead, only the node representation from the last layer is used to analyze the impact of multi-scale feature fusion on performance.

\begin{table*}[ht]
	\centering
	\caption{Node classification accuracy for ablation experiments on the Cora dataset. The best results in each column are highlighted in bold.}
	\label{tab:cora-ablation}
	\resizebox{\textwidth}{!}{%
		\begin{tabular}{lccccccccc}
			\hline
			\textbf{Method} & \textbf{5\%} & \textbf{15\%} & \textbf{25\%} & \textbf{35\%} & \textbf{45\%} & \textbf{55\%} & \textbf{65\%} & \textbf{75\%} & \textbf{85\%} \\
			\hline
			only1         & 0.7364 & 0.7980 & 0.8270 & 0.8389 & 0.8374 & 0.8424 & 0.8507 & 0.8516 & 0.8423 \\
			only2         & 0.7234 & 0.8015 & 0.8237 & 0.8368 & 0.8402 & 0.8369 & 0.8493 & 0.8516 & 0.8409 \\
			NoJK          & 0.7492 & 0.8048 & 0.8265 & 0.8393 & 0.8479 & 0.8487 & 0.8496 & 0.8595 & 0.8482 \\
			SDGNN-DGS0    & \textbf{0.7725} & \textbf{0.8232} & \textbf{0.8400} & \textbf{0.8480} & \textbf{0.8579} & \textbf{0.8620} & \textbf{0.8652} & \textbf{0.8673} & \textbf{0.8788} \\
			\hline
	\end{tabular}}%
\end{table*}

The results, as shown in Table~\ref{tab:cora-ablation}, indicate that removing any layer of aggregation or the Jumping Knowledge mechanism results in a performance decrease, with the removal of the first layer having the most significant impact, causing a nearly $0.04$ drop in accuracy at an 85\% training ratio. This demonstrates that the first layer's neighborhood community aggregation is crucial for capturing local structural patterns, while the second layer’s feature-driven clustering and cross-cluster fusion supplement high-order semantic information. Both layers working together ensure the best performance. At the same time, the Jumping Knowledge mechanism effectively alleviates the over-smoothing problem and improves classification performance by integrating multi-scale semantic information.

Combining the results of both parts, the best configuration for SDGNN is: two-layer structure $+$ two-stage aggregation (community partitioning and feature clustering) $+$ Jumping Knowledge mechanism. This combination ensures both classification accuracy and effective control over computational complexity, making it suitable for different scales and heterogeneous graph data scenarios.

\section{Conclusion}
\label{sec:conclusion}

This paper presents a parameter-free graph neural network framework, {SDGNN}, based on structural diversity, which models the structural heterogeneity and semantic consistency of node neighborhoods in a unified way without relying on any trainable parameters. The method uses the Structural-Diversity Message Passing mechanism, as the core computational paradigm. It combines three parameter-free group partitioning strategies and a structure enhancement mechanism based on normalized propagation, systematically characterizing the semantic interaction process of nodes in both local and global contexts. 

In the feature aggregation process, {SDGNN} not only enhances its ability to model structurally heterogeneous information through group partitioning, but also uses the normalized propagation matrix to achieve multi-hop structural–semantic fusion without training, significantly boosting the model’s representational ability and cross-domain generalization performance. Systematic experiments on eight public benchmark datasets and the interdisciplinary PubMed citation network demonstrate that {SDGNN} outperforms mainstream parameterized models, such as ComBSAGE, GCN, GAT, GraphSAGE, GIN, JK-Net, SGCN, and TANGNN, in terms of node classification performance, generalization ability, and robustness, particularly showing outstanding advantages in challenging scenarios such as low supervision, structural complexity, and sparse features. As a pioneering attempt, we anticipate that this study will motivate researchers to develop interpretable, high-performing, and more innovative graph neural networks grounded in classical sociological theories.




\bibliographystyle{elsarticle-num} 
\bibliography{bibliography}

\end{document}